\documentclass[10pt,twocolumn,letterpaper]{article}

\usepackage{iccv}
\usepackage{times}
\usepackage{epsfig}
\usepackage{subcaption}
\usepackage{graphicx}
\usepackage{amsmath}
\usepackage{amssymb}
\usepackage{amsfonts}

\usepackage{booktabs}
\usepackage{array,multirow}
\usepackage{float}
\usepackage{mathtools}

\usepackage[pagebackref=true,breaklinks=true,letterpaper=true,colorlinks,bookmarks=false]{hyperref}

\iccvfinalcopy % *** Uncomment this line for the final submission

\ificcvfinal\fi

\newcommand{\R}{\mathbb{R}}

\newcommand{\N}{\mathbb{N}}

\newcommand{\ra}[1]{\renewcommand{\arraystretch}{#1}}
\newcommand{\kraw}{K\textsuperscript{raw}}
\newcommand{\sraw}{S\textsuperscript{raw}}

\newcommand{\best}[1]{\textbf{#1}}

\newcommand{\dataimg}[1]{
    \begin{subfigure}{.23\textwidth}
        \centering
        \includegraphics[width=\linewidth]{#1}
        \caption{}
    \end{subfigure}%
}

\newcommand{\dataimglabelled}[2]{
    \begin{subfigure}{.23\textwidth}
        \centering
        \includegraphics[width=\linewidth]{#1}
        \caption{#2}
    \end{subfigure}%
}

\begin{document}

%%%%%%%%% TITLE
\title{Learning Optical Flow on ImageNet}
\title{Cheap Optical Flow Ground-Truth} % xD
\title{Inexpensive Method to Obtain Ground-Truth Optical Flow}
\title{Ground-Truth Is All You Need: An Inexpensive Method to Obtain Ground-Truth Optical Flow From Single Images}
\title{Optical Flow Ground-Truth Is All You Need}

% Learning Optical Flow from Superpixels
% It's time to let go of KITTI
% When Optical Flow Ground truth is scarce 
% 
\title{Learning Optical Flow from Single Images}  % this one is better than the previous
% as you prefer.
% yes let's stick with it
% can't think much ... so at the moment I have only one suggestion. Among the ones you provide I think the original
%is better. Then you choose what to use. Let's upload it.

\title{On Building Optical Flow Datasets from Single Images}
\title{An Approach to Build Optical Flow Datasets from Single Images}
\title{Optical Flow Dataset Synthesis from Unpaired Images}

\author{Adrian W\"alchli and Paolo Favaro\\
University of Bern, Switzerland\\
{\tt\small \{adrian.waelchli, paolo.favaro\}@inf.unibe.ch}
% For a paper whose authors are all at the same institution,
% omit the following lines up until the closing ``}''.
% Additional authors and addresses can be added with ``\and'',
% just like the second author.
% To save space, use either the email address or home page, not both
% \and
% Second Author\\
% Institution2\\
% First line of institution2 address\\
% {\tt\small secondauthor@i2.org}
}

\maketitle

%%%%%%%%% ABSTRACT
\begin{abstract}
   The estimation of optical flow is an ambiguous task due to the lack of correspondence at occlusions, shadows, reflections, lack of texture and changes in illumination over time. 
   Thus, unsupervised methods face major challenges as they need to tune complex cost functions with several terms designed to handle each of these sources of ambiguity. 
   In contrast, supervised methods avoid these challenges altogether by relying on explicit ground truth optical flow obtained directly from synthetic or real data. 
   In the case of synthetic data, the ground truth provides an exact and explicit description of what optical flow to assign to a given scene. 
   However, the domain gap between synthetic data and real data often limits the ability of a trained network to generalize. 
   In the case of real data, the ground truth is obtained through multiple sensors and additional data processing, which might introduce persistent errors and contaminate it. 
   As a solution to these issues, we introduce a novel method to build a training set of pseudo-real images that can be used to train optical flow in a supervised manner. 
   Our dataset uses two unpaired frames from real data and creates pairs of frames by simulating random warps, occlusions with super-pixels, shadows and illumination changes, and associates them to their corresponding exact optical flow. 
   We thus obtain the benefit of directly training on real data while having access to an exact ground truth. 
   Training with our datasets on the Sintel and KITTI benchmarks is straightforward and yields models on par or with state of the art performance compared to much more sophisticated training approaches.
\end{abstract}

%%%%%%%%% BODY TEXT
\section{Introduction}

% The awareness and processing of motion, visually or through other senses, is an essential skill that seemingly every animal needs for its survival.
% \emph{
%     Point to an impressive example of an animal that can react in nanoseconds to a predator attack through its visual system and point out how small its brain is
%     Provide a references how motion is processed from low level to high level in visual cortex.
%     While humans in the modern world do no longer depend on extreme survival skills, they develop and depend on new technologies that we expect to perform tasks a human can perform, faster, better, over a longer period of time.
%     An example is autonomous driving, a task that requires constant awareness of the state and motion of objects in the surrounding environments.
%     The optical flow, the apparent motion of pixels in series of video frames, is a long standing problem in Computer Vision that has yet to be solved.
% }
%While the field of Computer Vision has made great progress in image classification, segmentation etc through supervision, these advancements do not directly benefit the research in optical flow estimation as it is primarily a low-level task and the effort of manually annotating dense optical flow in only a couple of HD video frames is already comparable to annotating a full ImageNet~\cite{deng2009imagenet}. 

Optical flow estimation is the task of associating pixel locations across two consecutive frames in a video. More specifically, given two frames, the objective is to assign a shift at each pixel of the first frame, which corresponds to the location that that pixel has in the second frame. However, determining such correspondence is challenging due to ambiguities caused, for example, by occlusions, shadows, reflections, lack of texture and changes in illumination over time.

This task has been originally formulated as a per-image-pair energy optimization \cite{horn1981determining} and much of the effort has been devoted to designing regularization terms that would handle the ambiguities mentioned above and also capture some prior knowledge about the optical flow (\eg, smoothness and discontinuities). More recently, deep learning methods have allowed to train a single neural network model to estimate optical flow for any input image pair, with a remarkable improvement in terms of accuracy and estimation efficiency. The natural evolution of the original approach by Horn and Schunk~\cite{horn1981determining} has led to the unsupervised methods for optical flow \cite{janai2018unsupervised,jason2016back,liu2019ddflow,liu2019selflow,meister2017unflow,wang2018occlusion}. In fact, also these methods formulate a loss function for training that consists of a range of terms, each addressing one of the key ambiguities in the optical flow estimation task. The main difficulty in using and building on these methods is that they require a careful simultaneous tuning of several hyperparameters, and the number of these parameters grows with the number of challenges that we desire to address.

\begin{figure}[t]
    \centering
    \includegraphics[width=0.95\linewidth]{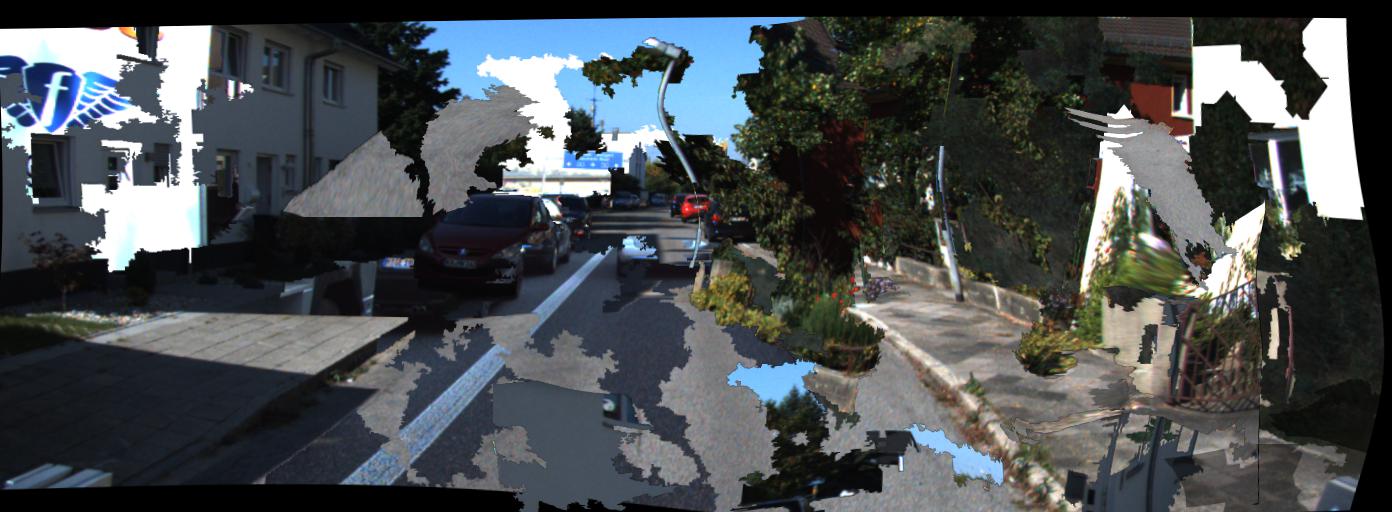}

    \vspace{1mm}
    
    \includegraphics[width=0.95\linewidth]{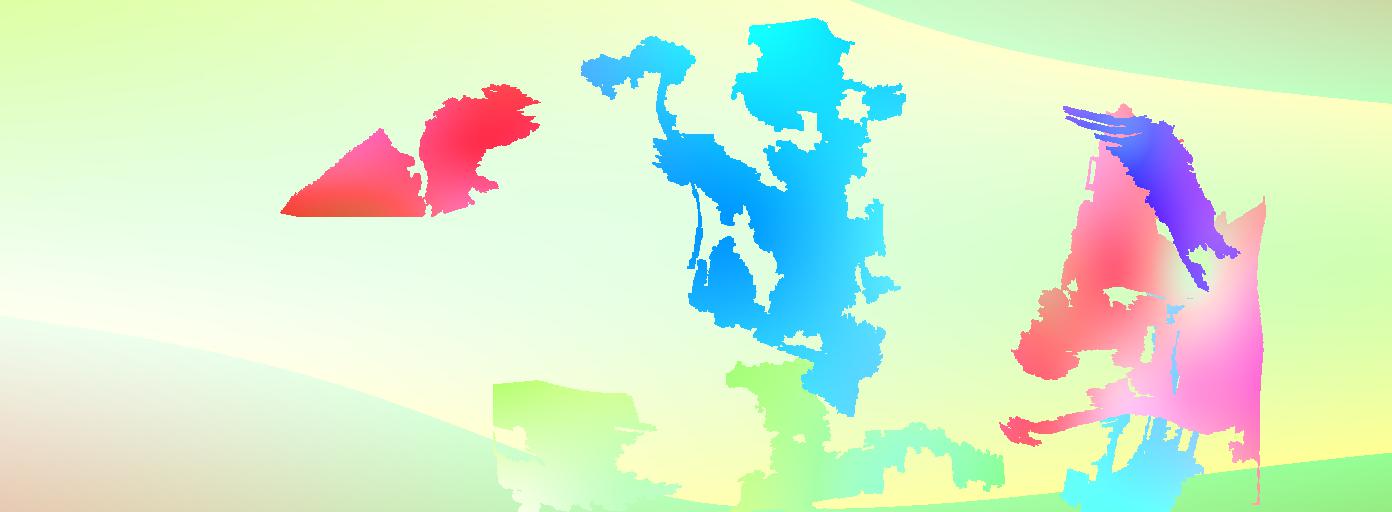}
    \caption{
        \label{fig:teaser}
        From a single image, we create a pair of frames and the corresponding exact optical flow ground truth by warping and displacing superpixels.
    }
\end{figure}

An alternative approach is supervised learning, which consists of training a network on data where the optical flow corresponding to image pairs is explicitly given. This solution moves the problem of designing and tuning multiple energy terms to the design of training data. Thanks to supervised learning the field of Computer Vision has made great progress in image classification, segmentation and object detection. However, while the manual annotation for ImageNet~\cite{deng2009imagenet} was expensive and time-consuming, but feasible, the manual definition of accurate ground truth optical flow for pairs of video frames is unmanageable and error-prone. One effort to automate the definition of ground truth optical flow is to use alternative sensors to measure it \cite{kitti}. However, real measures are often noisy and may be sparse. Thus, post-processing of measured optical flow is needed, but often introduces error bias. To avoid these limitations, synthetic datasets have been introduced, such as 
FlyingChairs~\cite{dosovitskiy2015flownet}, which consists of 3D chairs against a flat background texture, FlyingThings~\cite{flyingthings}, which uses a wider range of object classes than just chairs, and MPI-Sintel~\cite{sintel}, which is a 3D animated short film that comes with optical flow annotations directly extracted from the rendering pipeline.
% Due to the popularity of supervised learning early one, efforts were made to create artificial data from 3D renderings.
% There are several popular benchmark datasets that attempt to simulate realistic motions, and every year new methods emerge that set the state of the art performance on these datasets.
% State of the art supervised optical flow methods are trained with synthetic optical flow.
% Manual annotation of dense optical flow is an expensive and inaccurate process.
% For high resolution, high frame rate video it is unfeasible.
% FlyingChairs~\cite{dosovitskiy2015flownet} renders 3D chairs into a scene with a flat background texture.
% FlyingThings~\cite{flyingthings} is similar to FlyingChairs but has more object classes than just chairs.
% MPI-Sintel~\cite{sintel} is a 3D animated short film that comes with optical flow annotations directly extracted from the rendering pipeline.
Because these datasets are synthetic, the optical flow they provide is exact. However, synthetic and real data have also a domain gap and a network trained on one dataset might yield unpredictable outcomes on the other dataset.

%These datasets are often used for pretraining and benchmarking.
%While these datasets contain a diverse set of motions, they still involve hand crafting either the 3D shapes or their animation.
%\emph{Emphasize difficulties at occlusions, and ambiguity.}
In this work, we present a simple, flexible and inexpensive pipeline to generate as much optical flow ground truth data as needed and on the image domain the system is targeted for. Our data preparation requires no per-sample manual intervention and it only requires a set of (even non consecutive) images taken, for example, from a video or an unordered photo collection such as ImageNet \cite{deng2009imagenet}. The key idea is to take a single frame and to create two consecutive frames such that they share the texture of the given frame (and possibly some texture from another image), which provides both the foreground and the background. We first segment the single frame in superpixels, randomly choose a cluster of superpixels, apply a global translation and warping and then superimpose them to another part of the image. This operation is also repeated multiple times to simulate multiple occlusions. Moreover, the background (the unselected superpixels) are also warped in a random fashion. We show that this simple method for augmenting single images with optical flow ground truth is powerful enough to compete with state of the art methods that involve training on several synthetic and real datasets (see Figure~\ref{fig:teaser}).
%Learning from a single image: \cite{asano2020single}
% Can our simple warpings generalize to complex motions like rotations?
% When projected to 2D, rotations, scale and other deformation can be seen (locally) as a warping of the image texture.
% This simplification holds except for occlusions, which we handle separately. 
% Hence, if we can simulate image warpings, we don't need to simulate 3D motions of real objects in order to handle the same motions.

Our contributions are summarized as follows:
\begin{itemize}
    \item We introduce a general data pre-processing pipeline that is able to generate unlimited, diverse image pairs with ground truth optical flow in an unsupervised, inexpensive way;
    \item The data we generate is composed of realistic texture and occlusion boundaries, while the motions are synthetic;
    \item We show the role of texture in optical flow estimation and we find that our data generalizes best on the datasets used to build our synthetic training set.%We reduce the gap between the data distribution seen during training vs. the data seen at test time.
\end{itemize}

% 
% Differentiable superpixels end-to-end: \cite{jampani2018superpixel} 

\section{Prior Work}

\subsection{Supervised Flow Estimation}

The first end-to-end learning-based optical flow works 
\cite{dosovitskiy2015flownet, hui18liteflownet, ilg2017flownet} 
were inspired by the success of convolutional neural networks (CNNs) in per-pixel prediction tasks such as semantic segmentation \cite{long2015fully, ronneberger2015u} or depth estimation \cite{eigen2014depth}. 
FlowNet~\cite{dosovitskiy2015flownet} and all subsequent works take two or more consecutive video frames as input and estimate the dense forward flow between them.
These models are trained end-to-end with supervision through synthetic data and are fine-tuned on sparsely annotated real world videos~\cite{kitti}. 
Several works have made architectural improvements to boost performance, \eg, coarse-to-fine \cite{ranjan2017optical, sun2018pwc}, efficient cost volume computation \cite{sun2018pwc, yang2019volumetric} and warping in feature space~\cite{hui18liteflownet}.
Recently RAFT~\cite{teed2020raft} achieved state of the art performance with impressive numbers on all popular benchmarks. 
They introduce a new architecture based on GRU~\cite{cho2014learning} that performs progressive updates to a single flow estimate at multiple scales, and, in addition, the iterative approach enables an efficient correlation cost volume computation at full spatial resolution between all pairs of pixels.
Despite the impressive advancements in the supervised domain, these methods still require extensive pretraining with synthetic data \cite{dosovitskiy2015flownet, flyingthings} and need to be fine-tuned on sparse real-world data~\cite{kitti}, which is not readily available.
% We take their method as baseline and conduct all our experiments with their 

% \subsubsection{Optical Flow Ground Truth}

% FlyingChairs~\cite{dosovitskiy2015flownet} renders 3D chairs into a scene with a flat background texture.
% FlyingThings~\cite{flyingthings} is similar to FlyingChairs but has more object classes than just chairs.
% MPI-Sintel~\cite{sintel} is a 3D animated short film that comes with optical flow annotations directly extracted from the rendering pipeline.

\subsection{Unsupervised Flow Estimation}

%The lack of ground truth training data inspires the approaches that are unsupervised. 
Unsupervised methods try to avoid altogether the use of ground truth training data.
Early works focused on minimizing the photometric error between the first frame and the second frame warped by the estimated flow \cite{jason2016back}.
Together with a smoothing regularizer for the flow, this method is very effective in learning accurate predictions in non-occluded regions, but fails when the brightness-constancy constraint is not satisfied, \eg, at occlusion boundaries across specular surfaces.
Subsequent works improved these shortcomings by excluding the pixels in occluded regions from the loss using a mask obtained by forward warping \cite{wang2018occlusion} or a forward-backward consistency check~\cite{meister2017unflow}. 
Janai \etal~\cite{janai2018unsupervised} include multiple frames for occlusion reasoning to obtain sharper flow at boundaries.
Instead of excluding occluded pixels from the optimization, Liu \etal~\cite{liu2019ddflow} simulate the optical flow at image boundaries by cropping the images and imposing the flow predicted by a teacher model onto the student model.
\cite{liu2019selflow} take this idea a step further and hallucinate occlusions anywhere in the image with superpixels \cite{achanta2012slic}.
The advantage of these distillation methods is that the flow estimate for occluded target pixels can be included in the optimization objective, but they rely on a strong teacher model. 
Inspired by the distillation methods, ARFlow~\cite{liu2020learning} transforms (augments) a secondary pair of frames with a transformation and enforces the estimated optical flows to be consistent (equivariant) with the transformation.
% exploits equivariance as supervision signal by augmenting the pair of input frames with a transformation that must be consistent between the model output given the original pair and the model output given the transformed pair.
With only a handful of basic transformations such as cropping, zoom, affine and thin plate spline warps, and a model that has less than 2.5M trainable parameters, they achieve the state of the art performance among unsupervised methods. 

% \paragraph{} the state of the art among unsupervised methods, 
% They also apply spatial transformations (crop, flip, zoom,
% affine transform, thin-plate-spline), 
% appearance transformations (color jitter, brightness adjustment, blur, noise) and 
% occlusion transformations (random mask out on super pixels).

% In a similar fashion They initially train a model with photometric loss to obtain good predictions in non-occluded regions and continue training with hallucinated occlusions.

% Unsupervised methods require balancing several losses like photometric, smoothing, forward-backward consistency \cite{meister2017unflow}. 
% This can be challenging when not all terms can be satisfied at all times and contradict each other. 

% \paragraph{SelFlow\cite{liu2019selflow}}
% They initially train a model with photometric loss to obtain good predictions in non-occluded regions and continue training with hallucinated occlusions by selecting superpixels (cite SLIC) in the second image and replacing them with noise. 

% \subsubsection{Occlusion Hallucination}
% SelFlow
% Occlusions at boundaries: \cite{liu2019ddflow}

%\newpage

\section{Generating an Optical Flow Dataset}
%\section{Optical Flow Ground Truth from a Single Image}

%\subsection{Synthetic Occlusions}

\begin{figure}[tb]
    \dataimg{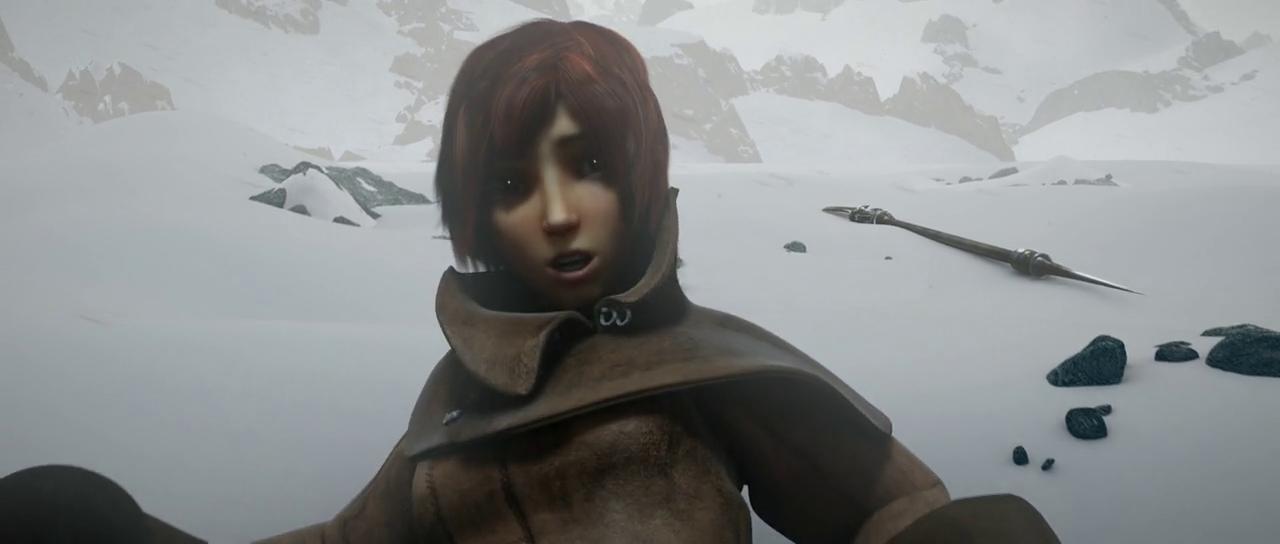}%
    \dataimglabelled{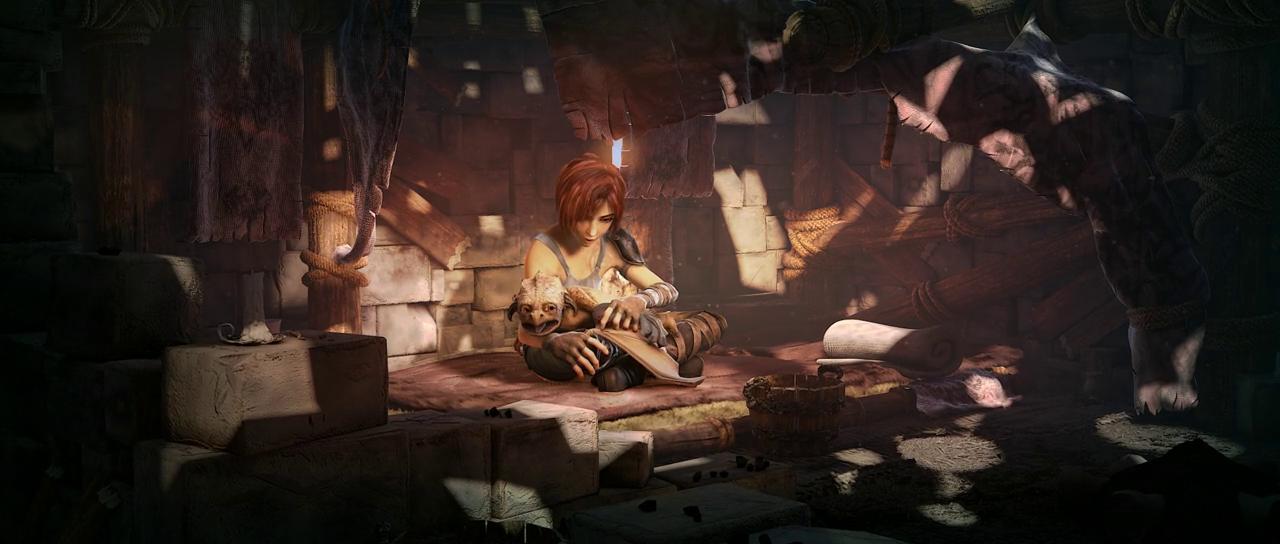}{\label{fig:aux-image}}\\
    \dataimg{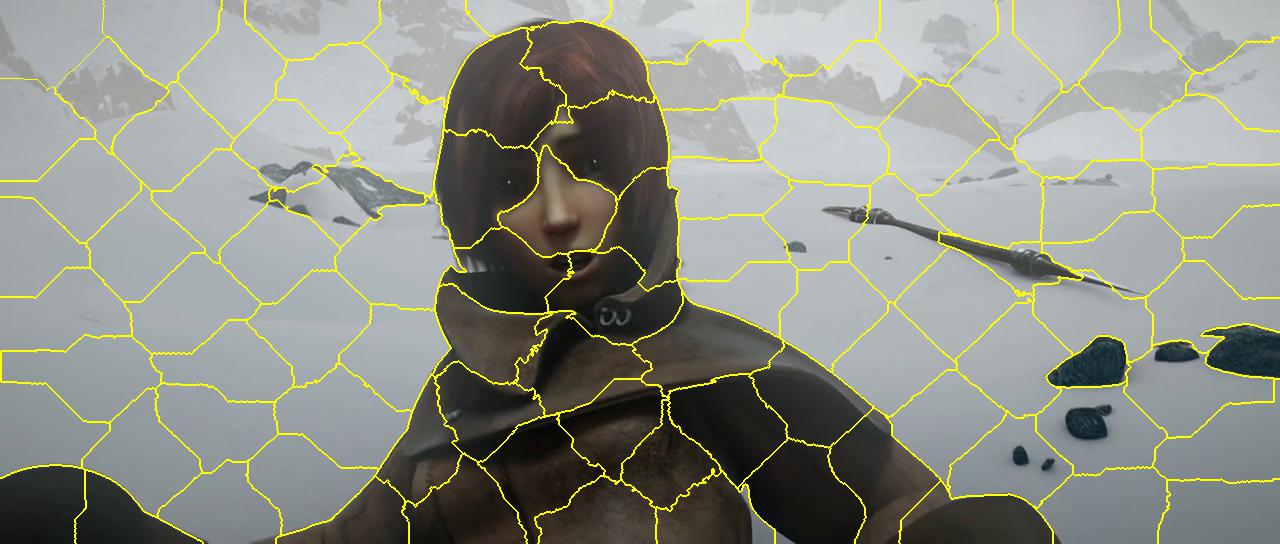}%
    \dataimg{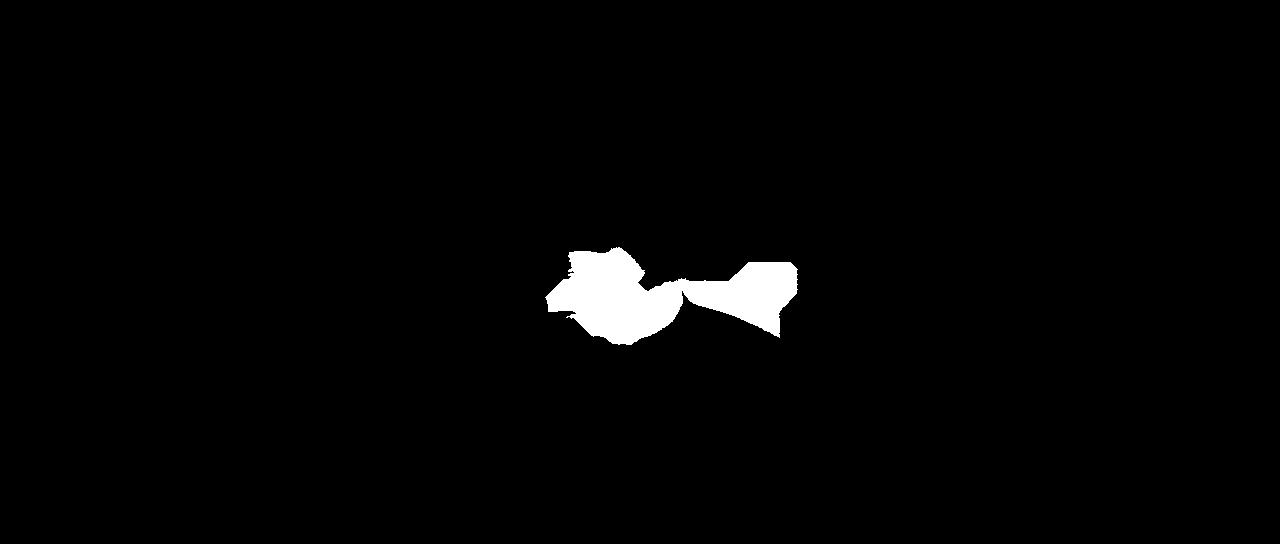}%
    \caption{
        \label{fig:occlusion-data-1}
        Superpixel extraction for occlusion simulation.
        (a) template image;
        (b) auxilliary image for background replacement;
        (c) superpixel segmentation of template;
        (d) selected superpixels for foreground motion.
    }
\end{figure}

Since our method follows the training procedure of supervised optical flow approaches, we present and discuss in detail only the preparation of the pseudo-synthetic dataset. For training, we use the state of the art neural network RAFT \cite{teed2020raft}.

\begin{figure}[tb]
    \dataimglabelled{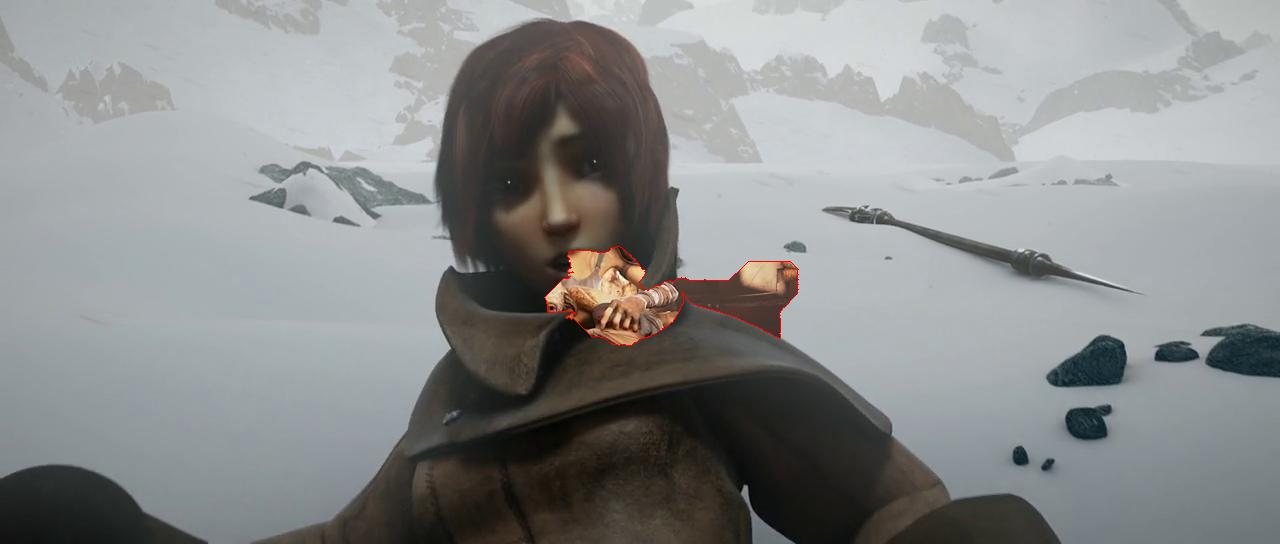}{\label{fig:bg-replaced}}%
    \dataimg{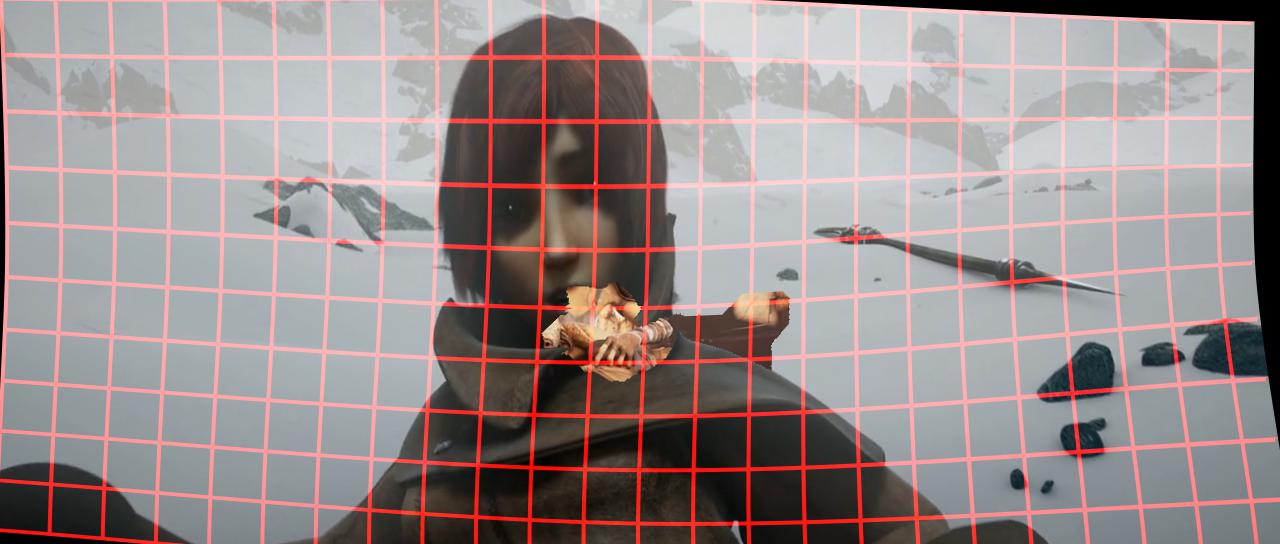}\\
    \dataimg{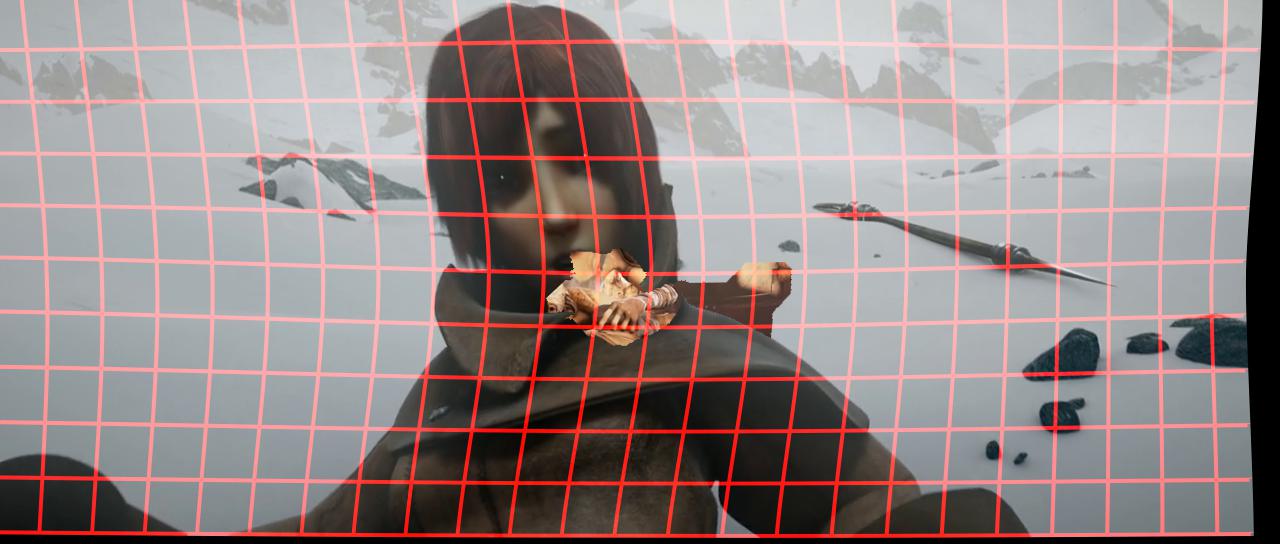}%
    \dataimg{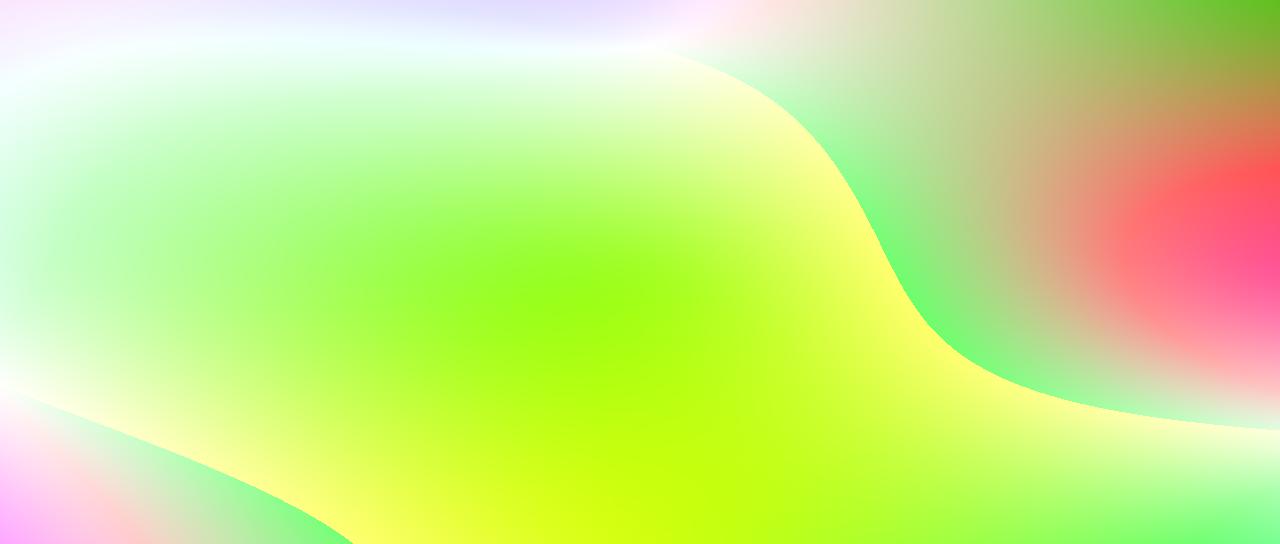}%
    \caption{
        \label{fig:occlusion-data-2}
        Background motion through warping.
        (a) Superpixel source texture replaced with auxiliary image;
        (b-c) Background motion through thin plate warping and corresponding forward flow (d).
    }
\end{figure}

\begin{figure}[tb]
 
    \dataimg{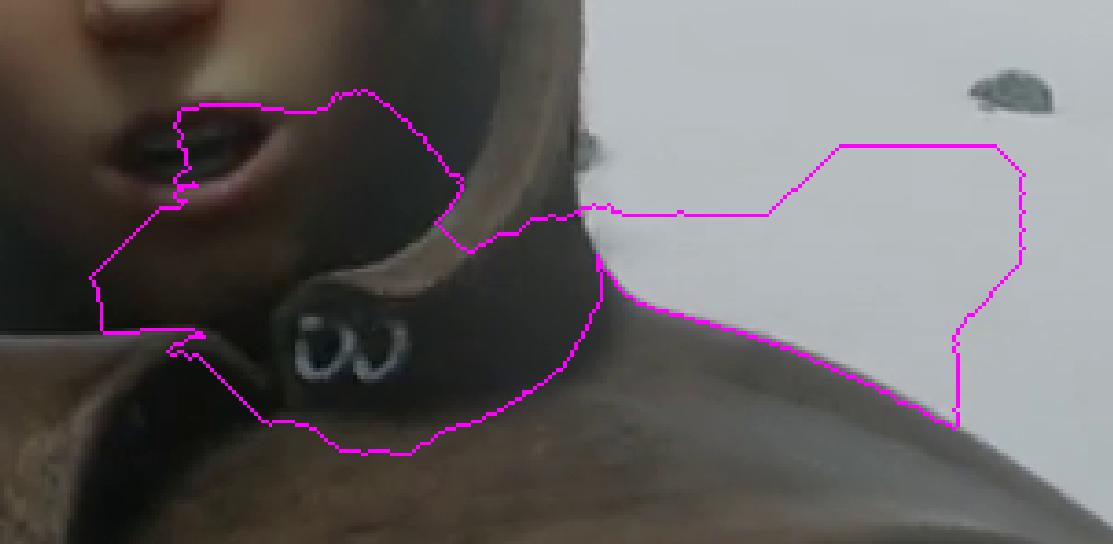}%
    \dataimg{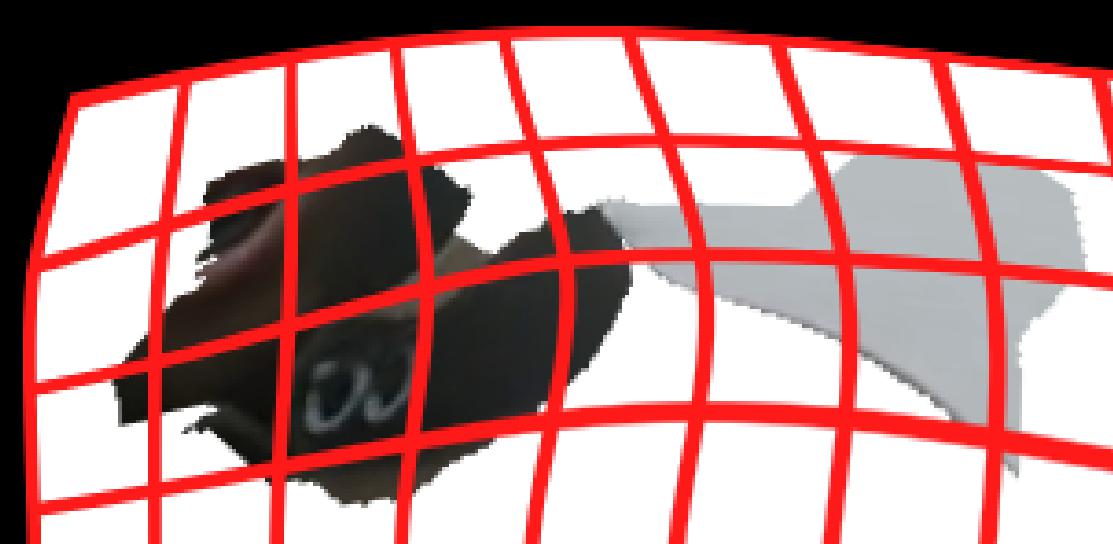}\\
    \dataimg{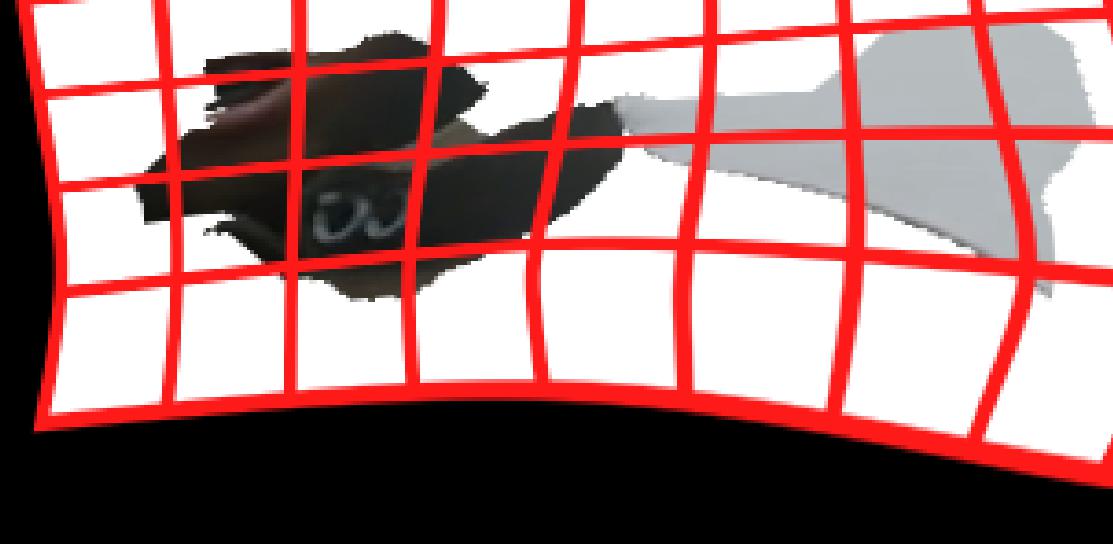}%
    \dataimg{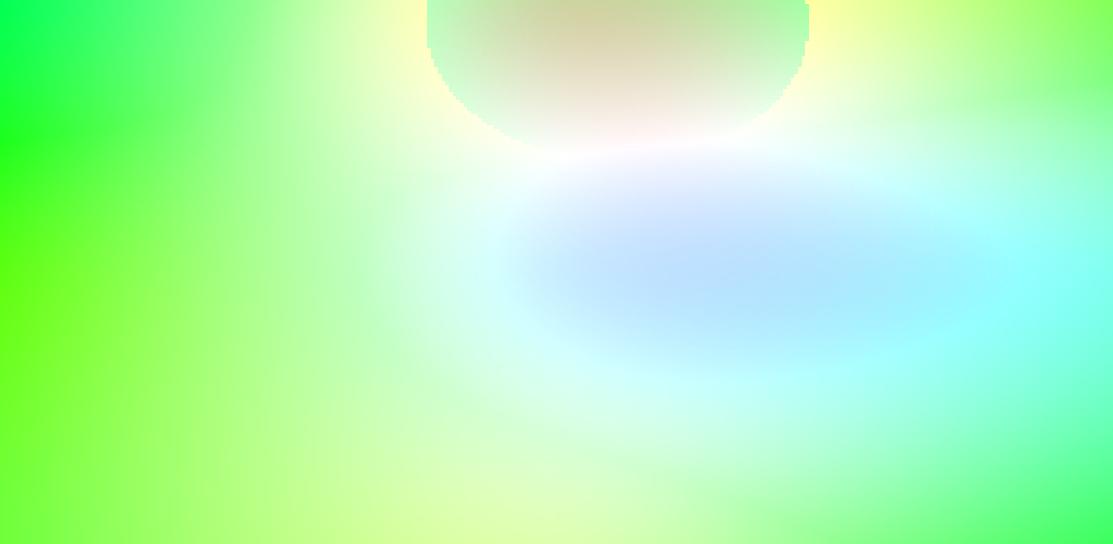}%

    \caption{
        \label{fig:occlusion-data-3}
        Foreground motion through warping.
        (a) Selected superpixel from original image in Figure~\ref{fig:occlusion-data-1}.
        (b-c) Foreground deformation through thin plate spline and corresponding optical flow (d).
    }
\end{figure}

\begin{figure}[tb]
    \dataimg{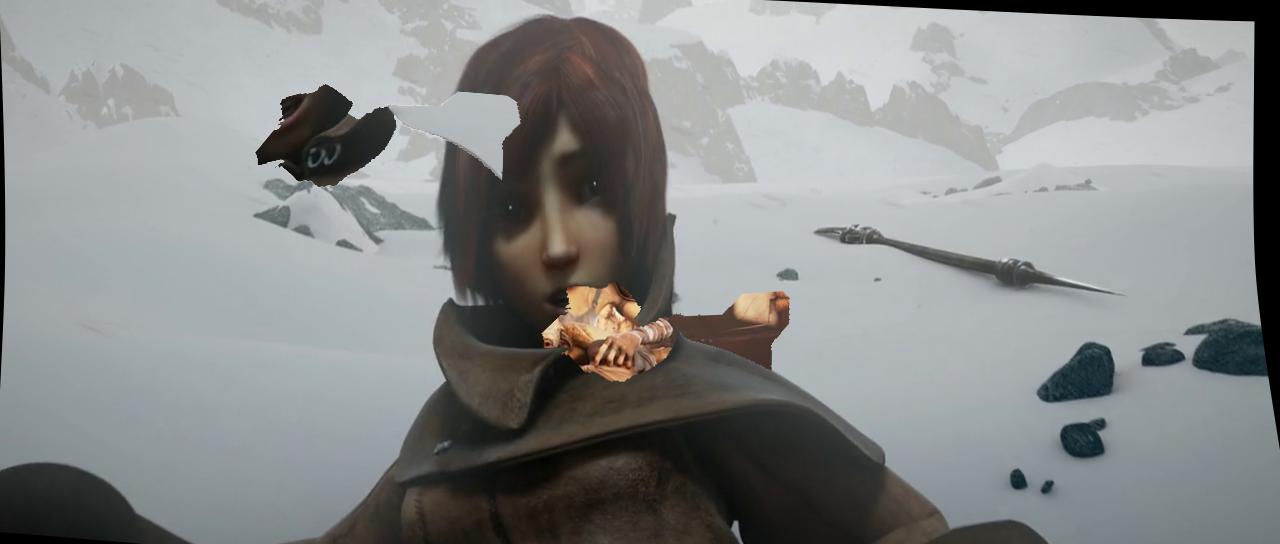}%
    \dataimg{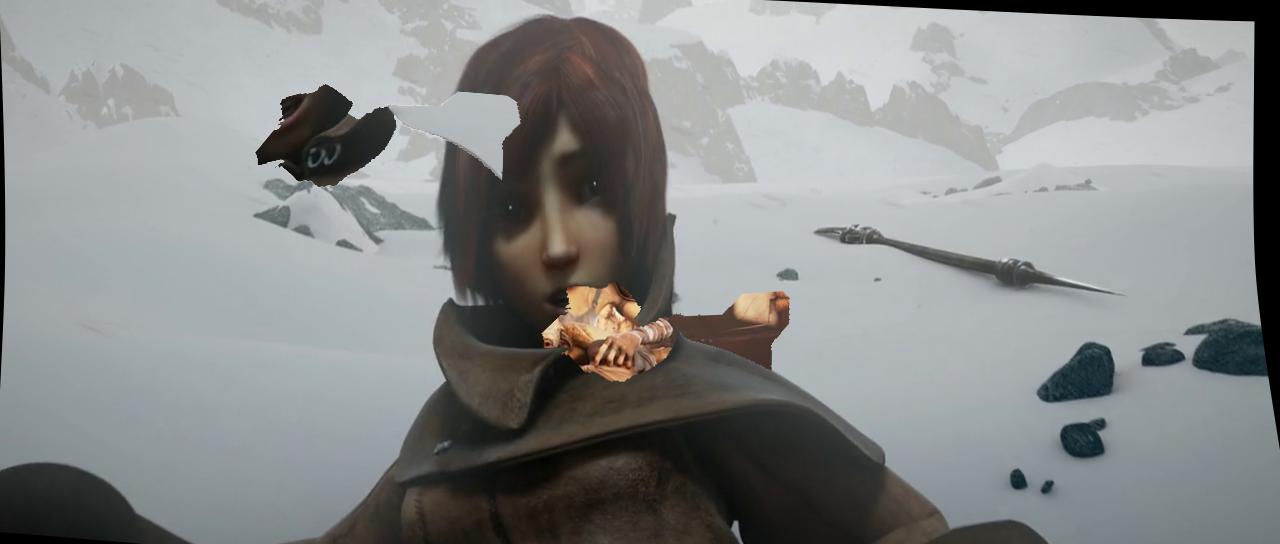}%
    \\
    \dataimg{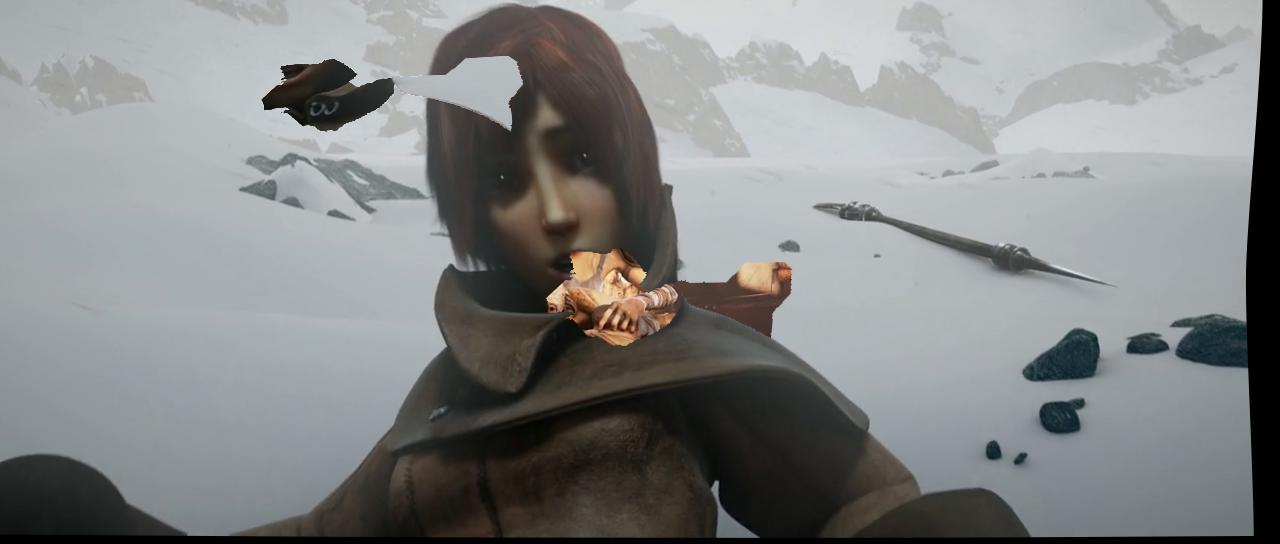}%
    \dataimg{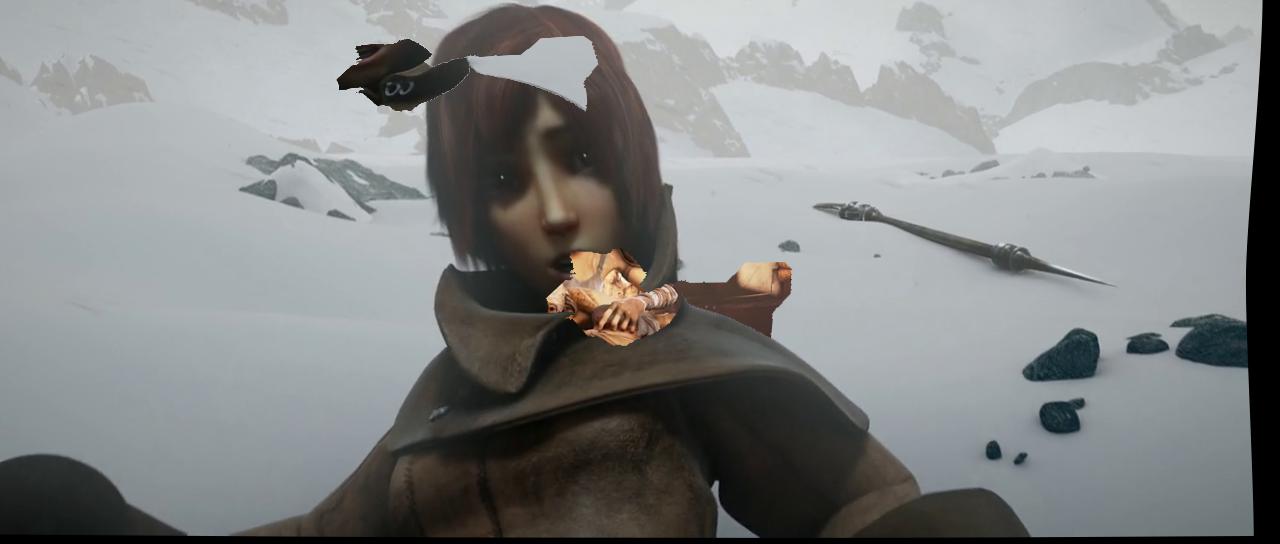}%
    \\
    \dataimg{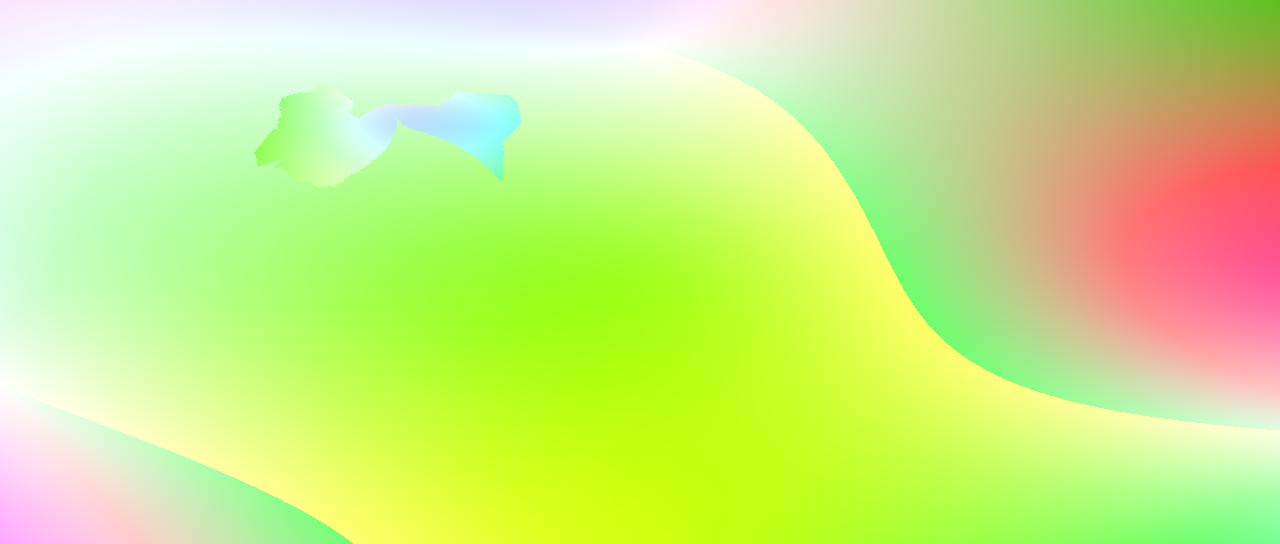}%
    \dataimg{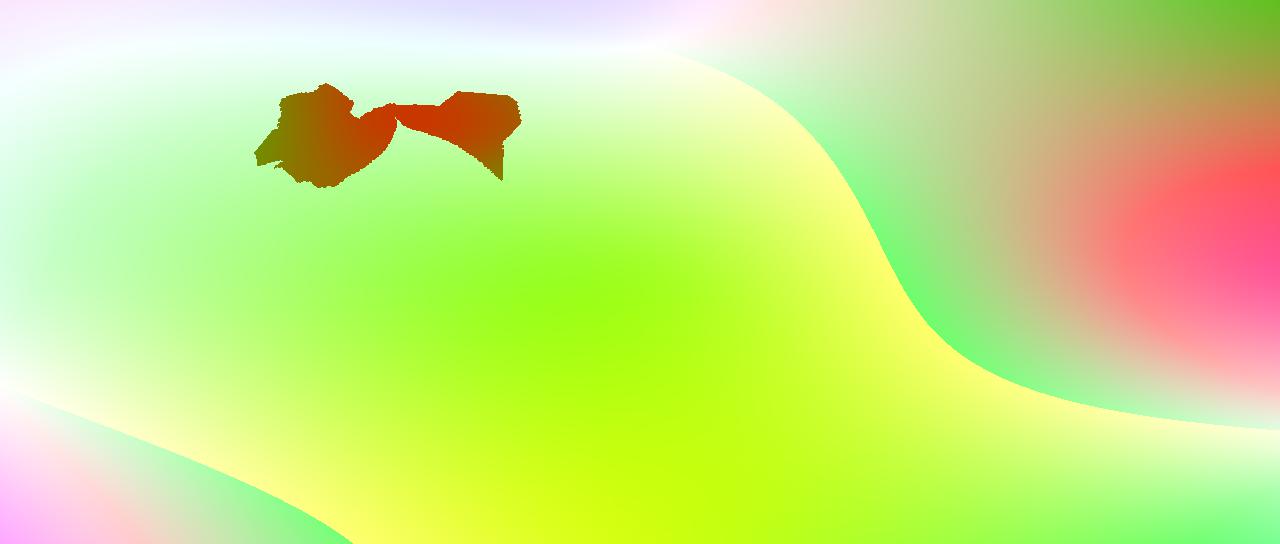}%
    \caption{
        \label{fig:occlusion-data-4}
        Composition of background and foreground motion.
        Left column (a,c,e): Background from Figure~\ref{fig:occlusion-data-2} with foreground from Figure~\ref{fig:occlusion-data-3} combined. 
        Right column: Additional global shift added to foreground.
    }
\end{figure}

We illustrate our data preparation pipeline in Figure~\ref{fig:occlusion-data-1}, \ref{fig:occlusion-data-2}, \ref{fig:occlusion-data-3} and \ref{fig:occlusion-data-4} using images from the Sintel~\cite{sintel} movie for clarity, though our method applies to all kind of image data. 
In the following paragraphs we describe our method of synthesizing motion using two random unrelated images.
The idea is to segment the image into parts and then shift and deform them independently on new layers to simulate occlusions and their corresponding exact optical flow.
% The idea is to segment and crop objects from an image and apply artificial warps and translations to
It is similar to what FlyingChairs~\cite{dosovitskiy2015flownet} and FlyingThings~\cite{flyingthings} do but less expensive and textures are specific to the testing domain.
In our method all textures for the foreground come from a single image and no additional assets like 3D geometry or shaders are required.

Given an image $I \in \R^{H \times W \times C}$, we compute superpixel segmentation maps $S_k \in \N^{H \times W}$ for $k = 1, \dots, K$ from coarse to fine, where $S_k(x) = i$ if the pixel $x$ belongs to superpixel $i$.
We use SLIC \cite{achanta2012slic} over other segmentation methods, because it is general, applicable to all domains and does not require pretraining.
Because all superpixels from a single scale $k$ are roughly the same size, we collect neighboring superpixels and group them together until a target size $m$ is reached. 
We denote the pixels that belong to the chosen group with a mask $M \in \{0, 1\}^{H \times W}$.
The superpixel segmentation and occluder mask $M$ is illustrated in Figure~\ref{fig:occlusion-data-1}.

Copying the pixels in $M$ to a different location in the same image would introduce an ambiguity in the motion between the new frames due to the duplication of texture.
We prevent this by painting the source region from where the superpixel was extracted with texture from an auxiliary image $A \in \R^{H \times W \times C}$ (see Figure~\ref{fig:aux-image}), \ie, 
\begin{equation}
    I^\prime = (1 - M) \odot I + M \odot A
\end{equation}
where $\odot$ is the element-wise multiplication.
The result $I^\prime$ is shown in Figure~\ref{fig:bg-replaced}.
Next, the single image $I^\prime$ is transformed twice with a Thin Plate Spline (TPS)  warping~\cite{bookstein1989principal, duchon1977splines} to simulate a smooth background motion between two frames. 
By applying the TPS transformation twice we make sure the two resulting images have an equal amount of distortion and interpolation artifacts. 
The extent of the deformation is limited by the \emph{control points}, which in our case are sampled by constructing a regular grid of size $L \times L$ and adding uniform noise to each grid point.
Applying the spline once to the regular grid of pixel coordinates $\{x_i\} \subset \R^{H \times W}$ yields a deformed grid $\{\phi(x_i)\}$ from which we can derive the per-pixel motion as $\phi(x_i) - x_i$.
Thus, we perform two image warpings with $\phi_1$ and $\phi_0$ in sequence, add a global shift $d$ to all coordinates and obtain a final background image pair
\begin{align}
    \label{eq:tps-warp-bg-1}
    B_1 (x) &= I^\prime (\phi_1(x)), \\
    \label{eq:tps-warp-bg-0}
    B_0 (x) &= B_1 (\phi_0(x) + d).
\end{align}
We use bilinear interpolation at non-integer locations.
The optical flow from image $B_0$ to image $B_1$ is
\begin{equation}
    \label{eq:tps-warp-bg-flow}
    \omega^{B}(x) = \phi_0(x) - x + d.
\end{equation}
% Given a number of control points $c_0, \dots c_l$ the TPS fitting yields a smooth function $f$
The pair $(B_0, B_1)$ and corresponding optical flow $\omega^B$ is shown in Figure~\ref{fig:occlusion-data-2}.

In the next stage, we extract the foreground texture $F = M \odot I$ from the original, undistorted image and apply two new TPS warps $\psi_1$ and $\psi_0$ analogous to eq.~\eqref{eq:tps-warp-bg-1} and~\eqref{eq:tps-warp-bg-0}
\begin{align}
    \label{eq:tps-warp-fg-1}
    F_1(x) &= F(\psi_1(x)) \\
    \label{eq:tps-warp-fg-0}
    F_0(x) &= F_1(\psi_0(x))
\end{align}
with the new foreground flow
\begin{equation}
    \label{eq:tps-warp-fg-flow}
    \omega^{F}(x) = \psi_0(x) - x.
\end{equation}
$F_0$ and $F_1$ with the corresponding optical flow is shown in Figure~\ref{fig:occlusion-data-3}.
The same warps $\psi_0$ and $\psi_1$ are applied to the mask $M$ and we get two new masks $M_0$ and $M_1$.

Finally, we combine the foreground with the background by placing the foreground $F_0$ and $F_1$ at positions $p_0$ and $p_1$, respectively, \ie,    
\begin{multline}
    \label{eq:fg-placement}
    I_i(x) = M_i(x - p_i) F_i(x - p_i) + (1 - M_i(x - p_i)) B_i(x), 
\end{multline}
for $i = 1,2$ and $\delta = p_1 - p_0$ is the relative global motion of the foreground from the first to the second image.
The optical flow is composed by 
\begin{equation}
    \label{eq:fg-flow-placement}
    \omega(x) = M_0(x - p_0) \omega^F(x - p_0) + (1 - M_0(x - p_0)) \omega^B(x). 
\end{equation}
In Figure~\ref{fig:occlusion-data-4} we show an example with $p_0 = p_1$, \ie no global foreground motion ($|\delta|= 0$) and an example with $|\delta| > 0$.

\begin{figure*}
    % sample 1
    \includegraphics[width=0.33\linewidth]{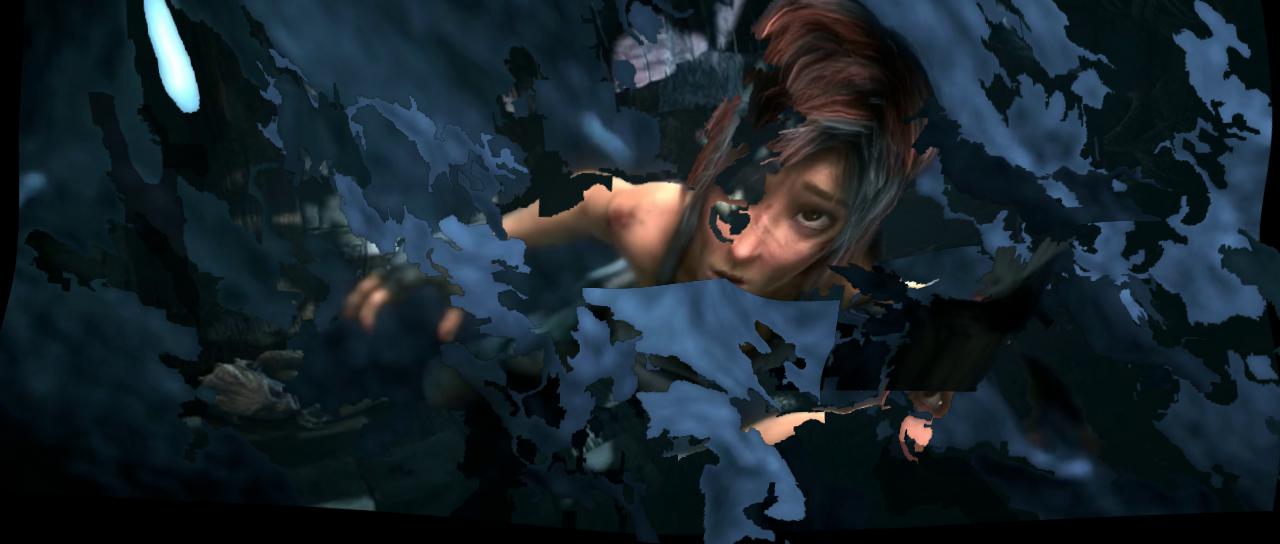}
    \includegraphics[width=0.33\linewidth]{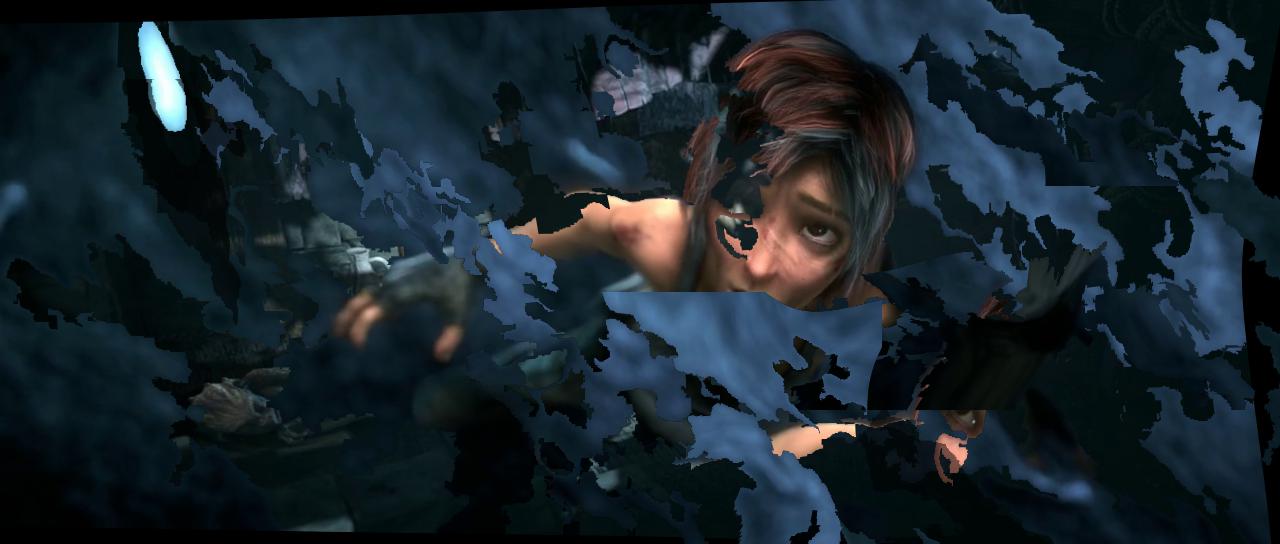}
    \includegraphics[width=0.33\linewidth]{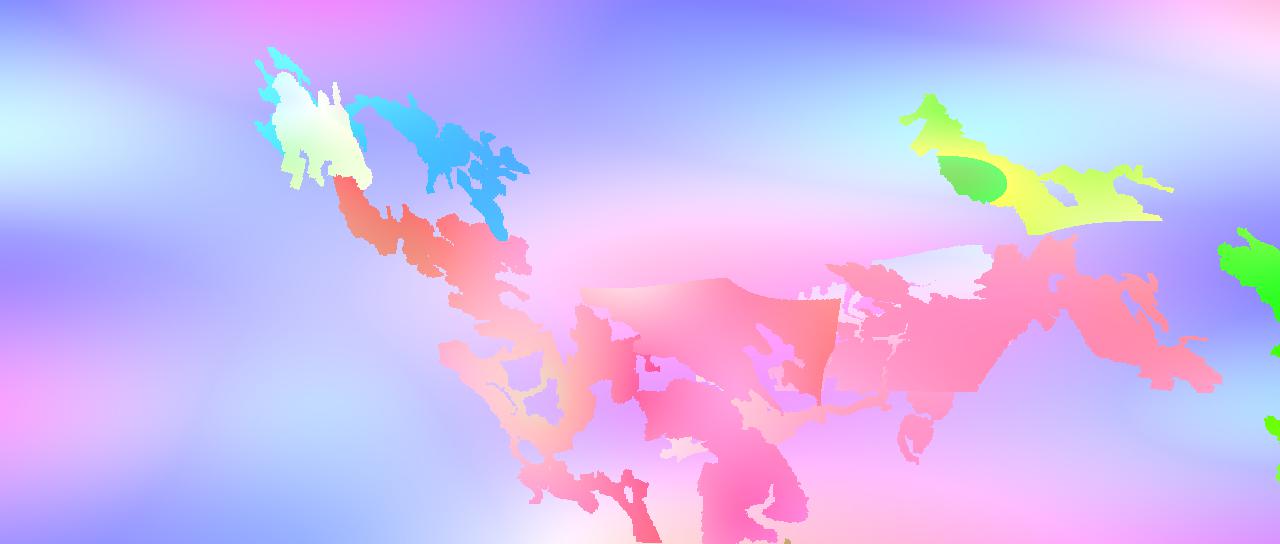}\\
    % sample 2
    \includegraphics[width=0.33\linewidth]{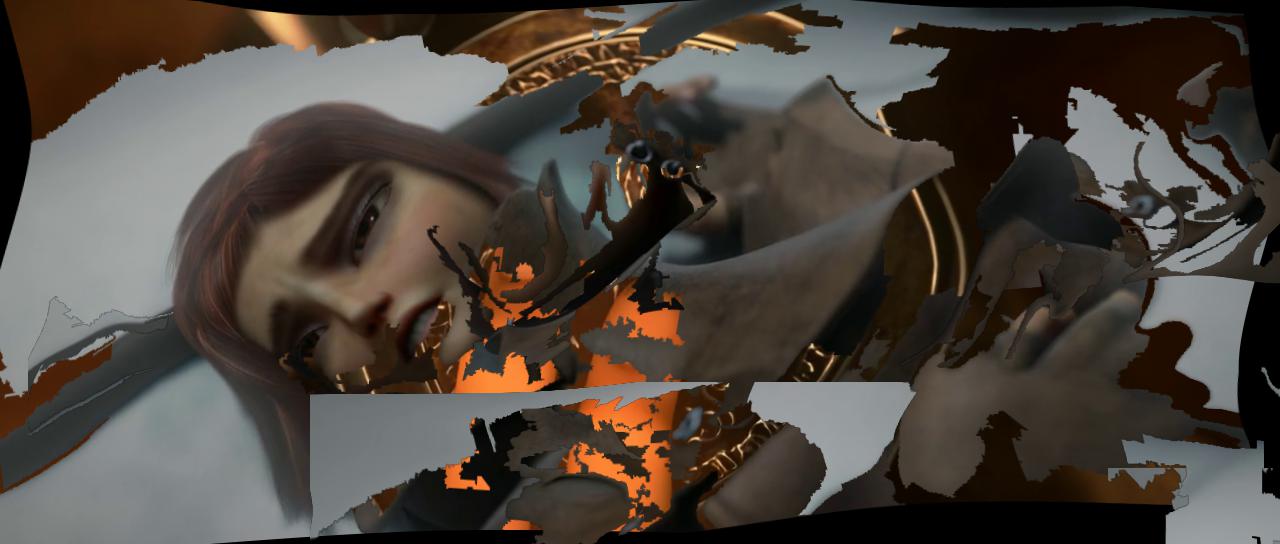}
    \includegraphics[width=0.33\linewidth]{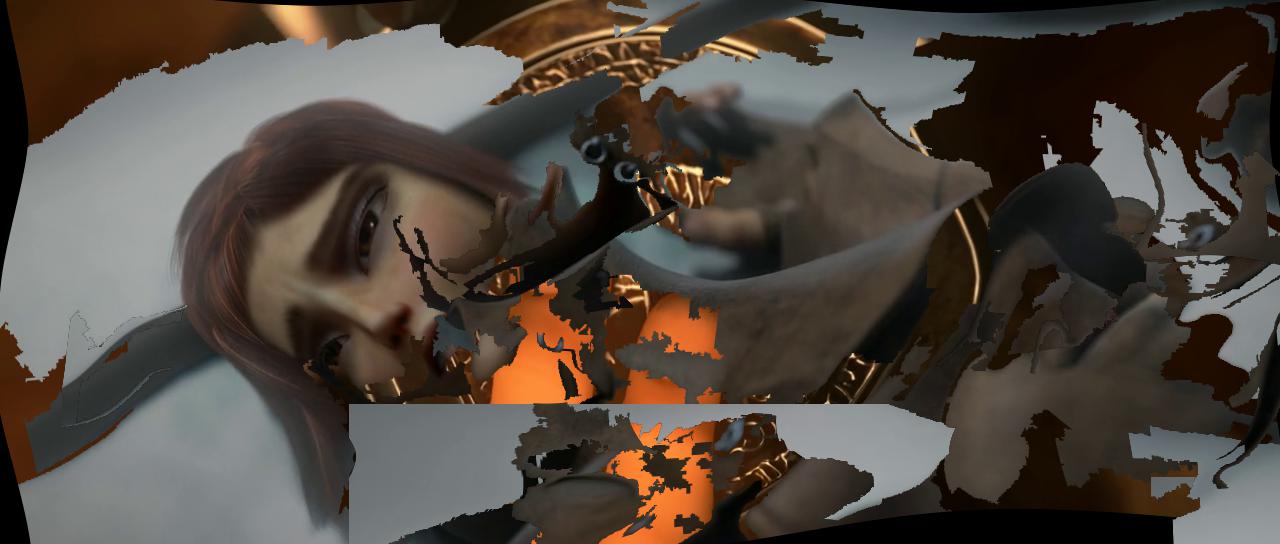}
    \includegraphics[width=0.33\linewidth]{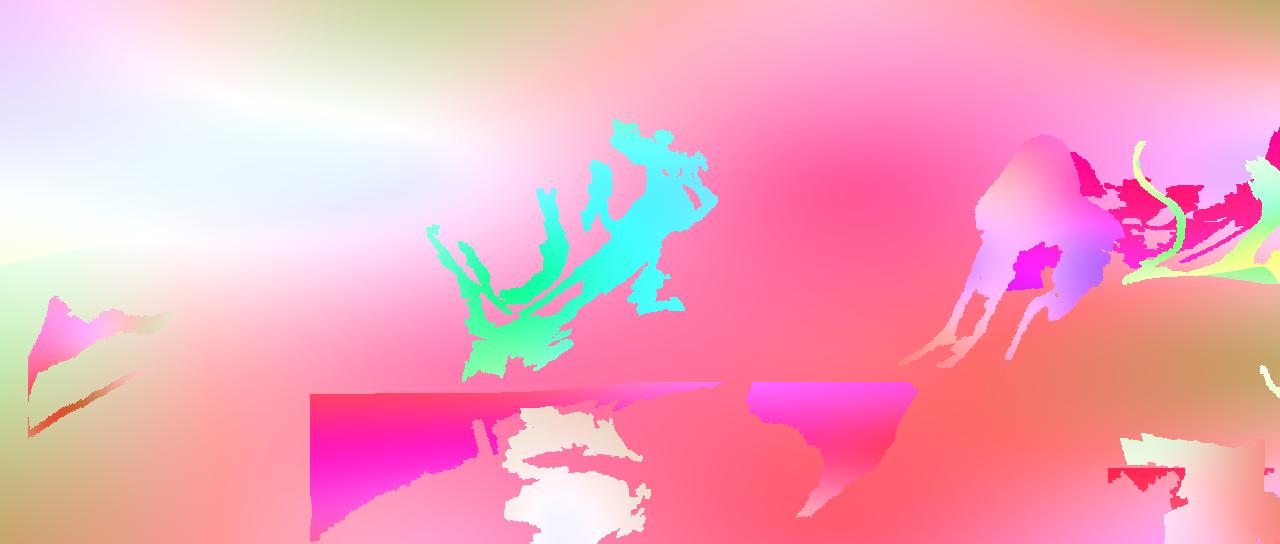}\\
    % sample 1
    \includegraphics[width=0.33\linewidth]{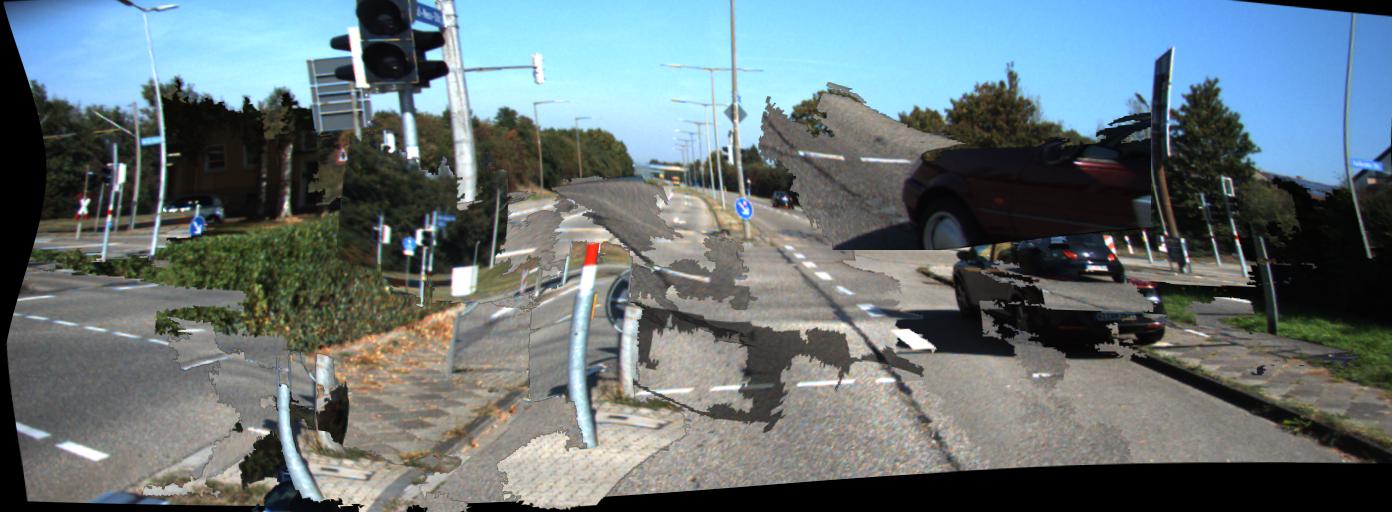}
    \includegraphics[width=0.33\linewidth]{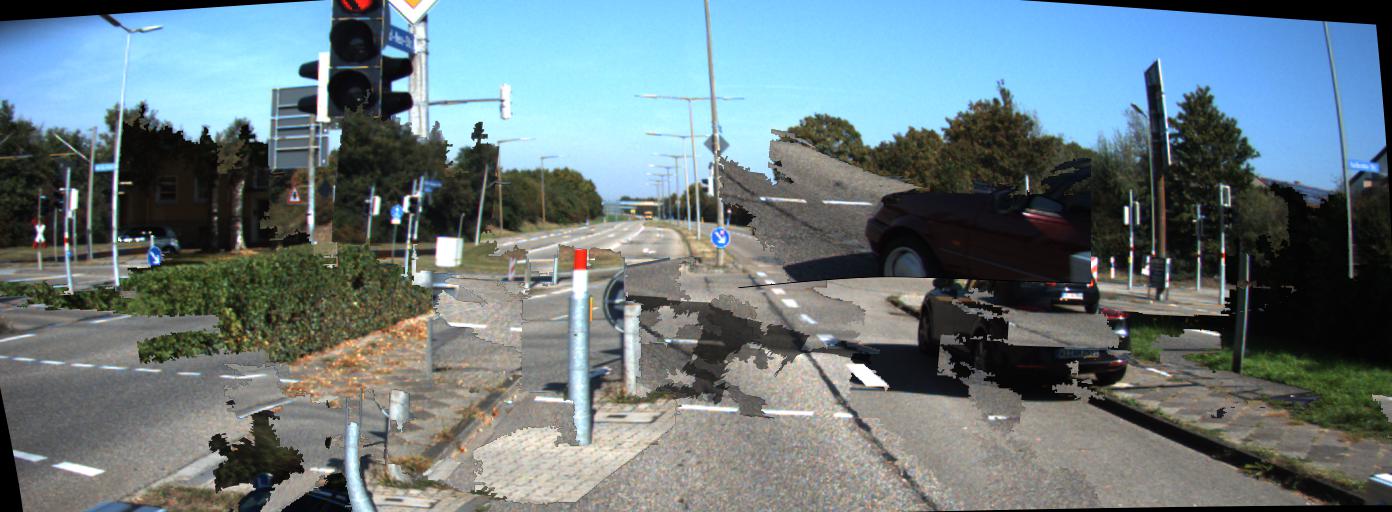}
    \includegraphics[width=0.33\linewidth]{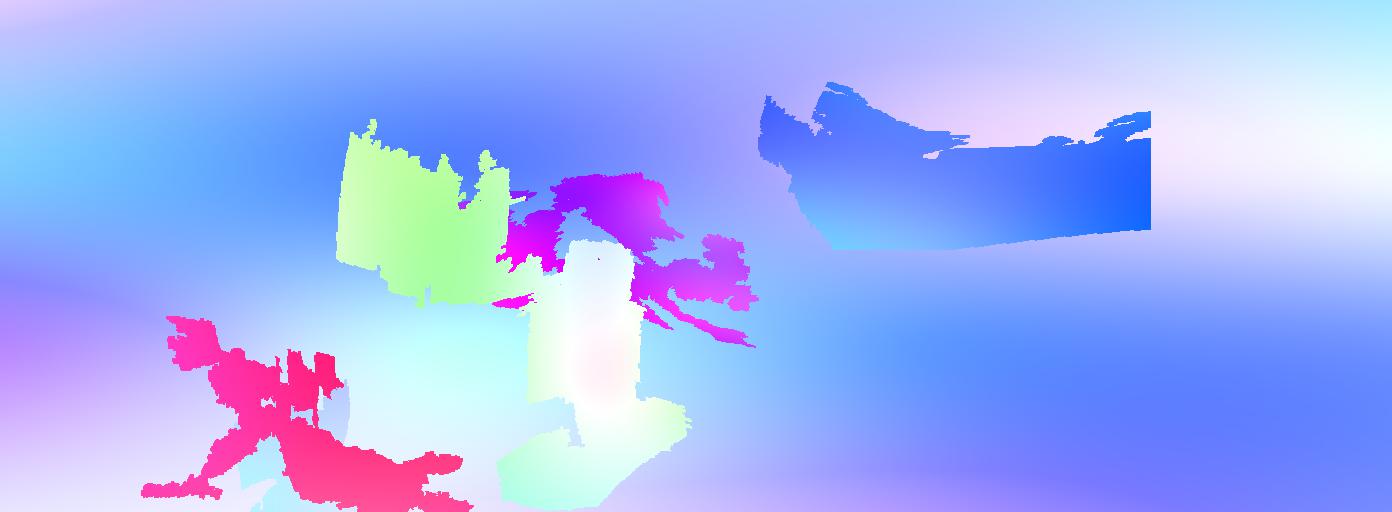} \\
    % sample 2
    \includegraphics[width=0.33\linewidth]{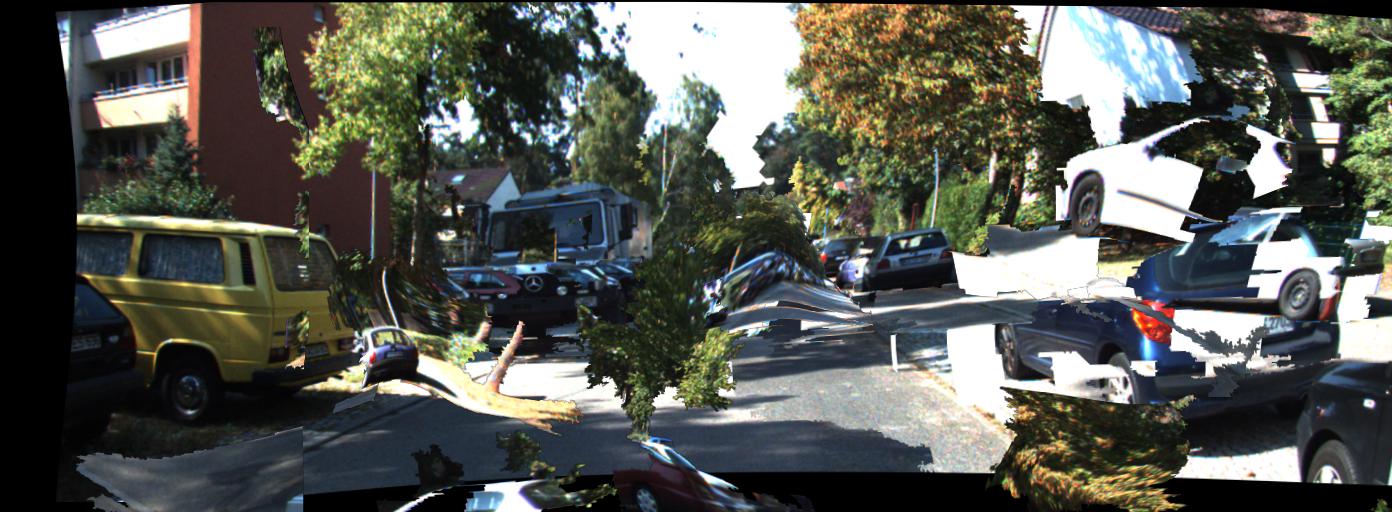}
    \includegraphics[width=0.33\linewidth]{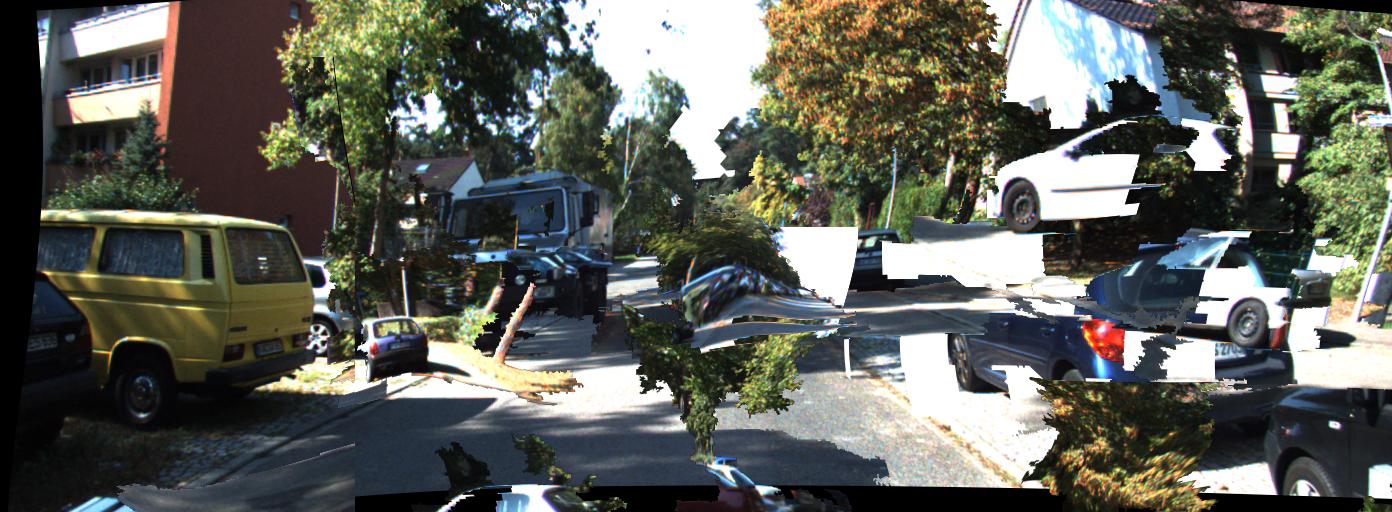}
    \includegraphics[width=0.33\linewidth]{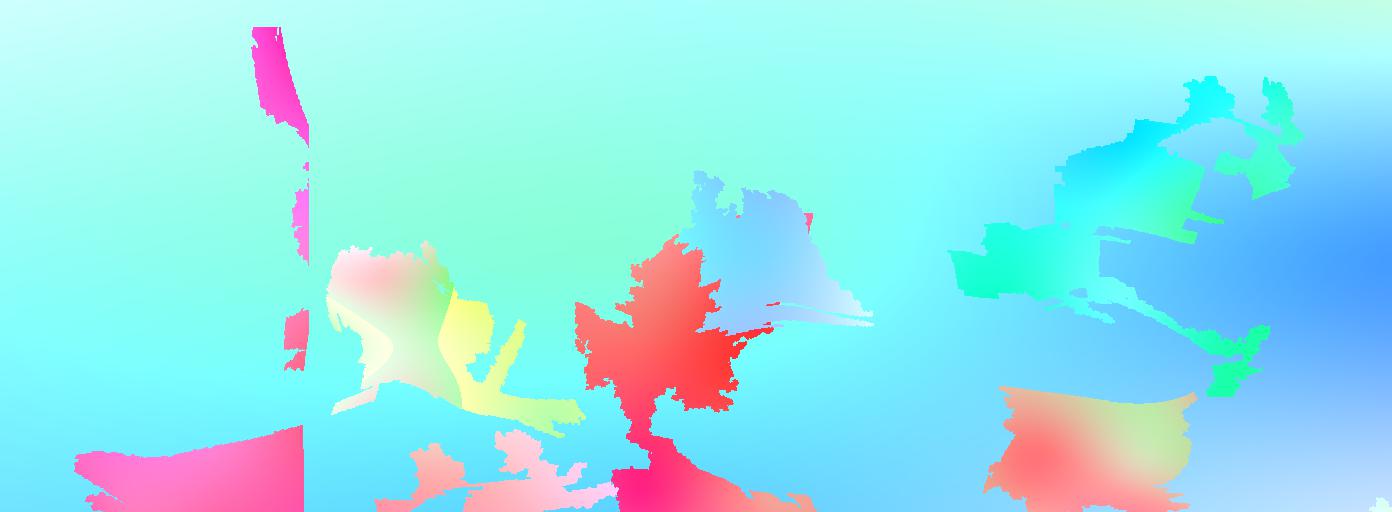} \\
    \caption{
        \label{fig:data-examples}
        Examples of our training data (first frame, second frame and optical flow) built on either Sintel~\cite{sintel} (top) or KITTI~\cite{kitti} (bottom).
        More examples can be found in the supplementary material.
    }
\end{figure*}

The above procedure can easily be adapted to generate a pair of frames with several (overlapping) occlusions.
Instead of choosing a single group of superpixels, we sample $N$ distinct groups and each time draw $k$ and $m$ from a uniform distribution, yielding individual mask $M_1, \dots, M_N$.
All masked regions get inpainted collectively to form a single background and are then warped using eq.~\eqref{eq:tps-warp-bg-1}, \eqref{eq:tps-warp-bg-0} and \eqref{eq:tps-warp-bg-flow}.
The different foreground superpixels and their motions get added in order of depth following eq.~\eqref{eq:tps-warp-fg-1} - \eqref{eq:fg-flow-placement}, with the last layer on top.
Example data is shown in Figure~\ref{fig:data-examples}.

With our simple occlusion synthesis, we are able to impose the exact optical flow in ambiguous regions where a high-level of reasoning based on the environment is required to recover the motion, \eg, at the occlusion boundaries or on surfaces with uniform or repeated textures. 

\subsection{Additional Augmentations}
We add additional augmentations as a means to increase the amount of data, and also for the purpose of robustness.
With probability $p_\text{s}$, we convert the foreground layer to a shadow, which we model as a semi-transparent black texture with the shape of the superpixel.
All warpings and translations remain the same, except that we set $\omega^F \coloneqq \omega^B$ in eq.~\eqref{eq:fg-flow-placement}, \ie, we impose invariance to transparent objects and their motion. 
Following prior works \cite{ilg2017flownet, teed2020raft}, we add color jittering to simulate global illumination changes, scaling transforms to both views, random horizontal and vertical flips, random crop, and random erasing in a rectangle in the second view. 
The optical flow is transformed as well according to the augmentation performed on the images.

\section{Experiments}

We conduct all experiments on the standard benchmark datasets Sintel and KITTI and take the raw, unlabelled data for our unsupervised data preprocessing.
Quantitative performance is measured by the average End-Point Error (EPE, Euclidean distance between predicted- and ground truth flow) and the F1-all outlier ratio (percentage of EPE that are larger than 3 pixels or not within 5\% of the ground truth).
We use the state of the art RAFT \cite{teed2020raft} as a baseline and train with their code and exact settings for fair comparison.
The only difference is the data.

\paragraph{Implementation.}
We precompute two superpixel segmentation maps ($K = 2$) per image with 100 and 1000 components.
In each image, we select between $N = 8$ and $N = 14$ superpixel groups for foreground motion.
Each superpixel group is chosen according to the maximum size $M \sim \mathcal{U}\{6000, 50000\}$ and for each deformation (foreground and background) we sample the grid size of TPS control points $L \sim \mathcal{U}\{3, 5\}$.
The auxiliary image is randomly picked from the whole dataset.
For the foreground and background motion, we limit the extent of the warping by sampling the displacement of the individual control points from the normal distribution $\mathcal{N}(0, 25^2)$.
Similarly, the global background translations $d$ and the foreground translations $\delta$ are sampled from $\mathcal{N}(0, 30^2)$. 
The probability of a foreground becoming a shadow is $p_\text{s} = 0.2$ and the transparency is also chosen randomly according to $\mathcal{U}(0.4, 0.6)$.
Our method is implemented in \mbox{PyTorch}~\cite{falcon2019pytorch, paszke2017automatic}. 
Code and data are available online.\footnote{Code will be released upon publication.}

\begin{table*}[t]
   %  \footnotesize	
    \ra{1.3}
    \centering 
    \begin{tabular}{c|lccccccccc}
        \toprule
            \multirow{2}{*}{Data} &
            \multirow{2}{*}{Method} & 
            \multicolumn{2}{c}{SINTEL (train)} & 
            \multicolumn{2}{c}{SINTEL (test)} &
            \multicolumn{2}{c}{KITTI 12} & 
            % \multicolumn{1}{c}{KITTI 15 (train)} &
            \multicolumn{2}{c}{KITTI 15} &
            
        \\
        % \cmidrule{2-3} \cmidrule{5-6} \cmidrule{8-9}
            & & clean & final & clean & final & train & test & train & test (F1)
        \\ 
        \midrule
        SY / \kraw & UnFlow-CSS~\cite{meister2017unflow}
        & - & (7.91) & - & 10.22 & 3.29 & - & 8.10 & -
        \\
        C + S / K & OccAwareFlow~\cite{wang2018occlusion}
        & (4.03) & (5.95) & 7.95 & 9.15 & - & - & - & -
        \\
        R + \sraw / \kraw & MFOccFlow\textsuperscript{$\dagger$}~\cite{janai2018unsupervised}
        & (3.89) & (5.52) & 7.23 & 8.81 &  - & - & - & -
        \\
        \kraw & BridgeDepthFlow\textsuperscript{\S}~\cite{lai2019bridging}
        & - & - & - & - & 2.56 & - &  7.02 & -
        \\
        \kraw & CCFlow\textsuperscript{\S}~\cite{ranjan2019competitive}
        & - & - & - & - & - & - & 5.66 & 25.27\%
        \\
        \kraw & UnOS-stereo\textsuperscript{\S}~\cite{wang2019unos}
        & - & - & - & - & 1.64 & 1.80 &  5.58 & 18.00\%
        \\
        % & & EpiFlow-train-ft\textsuperscript{\S} 
        % & (3.54) & (4.99) & (2.51) & - &  (5.55) & -
        % \\
        C + S / K & DDFlow~\cite{liu2019ddflow}
        & (2.92) & (3.98) & 6.18 & 7.40 & 2.35 & 3.00 & 5.72 & 14.29\%
        \\
        S + \sraw / K & SelFlow\textsuperscript{$\dagger$}~\cite{liu2019selflow}
        & (2.88) & (3.87) & 6.56 & 6.57 & 1.69 & 2.20 & 4.84 & 14.19\%
        \\
        \sraw / \kraw & ARFlow~\cite{liu2020learning}
        & 2.79 & 3.73 & 4.78 & 5.89 & 1.44 & 1.80 & 2.85 & 11.80\%
        \\
        \midrule
        \sraw / \kraw & \textbf{RAFT-Ours} ``C''
        & 2.00 & 3.49 & - & - & 2.42 & - & 6.17 & - 
        \\
        \sraw / \kraw & \textbf{RAFT-Ours} ``C + T''  
        & \best{1.93} & \best{3.47} & \best{4.06} & \underline{5.97} & 2.39 & 2.90 & 5.99 & 15.75\%
        \\
        \kraw / \sraw & \textbf{RAFT-Ours} ``C''
        & 2.60 & 4.15  & - & - & 2.76 & - & 6.41  & -
        \\
        \kraw / \sraw & \textbf{RAFT-Ours} ``C + T''  
        & 2.14 & 3.82 & - & - & 2.57 & - & 6.66 & -
        \\
        \bottomrule
    \end{tabular}
    \caption{
        \label{tbl:comparison-unsupervised}
        \textbf{Comparison with unsupervised methods} on MPI-Sintel~\cite{sintel} and KITTI~\cite{kitti} benchmarks.
        All numbers are AEPE (average end-point-error) except for the KITTI test set where outlier ratio F1 is reported.
        Numbers enclosed by parentheses mean that the training is performed on the same dataset.
        The results for entries marked as (-) were not reported by the original authors.
        Methods marked with $\dagger$ are using more than two frames.
        Methods trained with geometrical constraints are marked with \S.
    }
\end{table*}

\paragraph{Sintel.} This is an animated movie with $\sim$21k video frames from which a subset of $\sim$1k pairs are annotated with ground truth optical flow and occlusion masks. 
There exist two versions, a clean and final pass, where the latter contains more realistic motion blur and illumination.
The clean and final subsets are only used for evaluation.
% original size 1024 × 436 pixel
We download the raw version of the movie rendered at a resolution of $1280 \times 544$ pixels and manually discard roughly 1.6k frames that belong to scene transitions, intro and credits roll.
We label this dataset as S\textsuperscript{raw} in our tables.

\paragraph{KITTI.} 
The KITTI dataset contains a large amount of raw stereo image data. 
There are a total of $\sim$49k images with dimensions $1242 \times 375$ pixels.
We take all images from the left view and apply our unsupervised data augmentation to all video frames.
We refer to this dataset as K\textsuperscript{raw} in our tables.
The evaluation is done on the official training and test splits of the KITTI data (denoted as K), where the training split has sparse optical flow labels.
Samples of our KITTI data are shown in Figure~\ref{fig:data-examples}.

\paragraph{Comparison with Unsupervised Methods.}
Since our data is generated in an unsupervised way, we compare against other unsupervised methods in Table~\ref{tbl:comparison-unsupervised}. 
RAFT~\cite{teed2020raft} was originally published as a supervised method trained in stages on FlyingChairs~\cite{dosovitskiy2015flownet}, FlyingThings~\cite{flyingthings}, Sintel~\cite{sintel} and KITTI~\cite{kitti} in this order for 100k iterations per stage.
We retrain their model with our data following the same training schedules and augmentation settings and we evaluate after each stage.
The results are shown in Table~\ref{tbl:comparison-unsupervised}. 
We label our models evaluated after the FlyingChairs stage as ``C'' and the models evaluated at the FlyingThings stage with \mbox{``C + T''}. 
The results show that RAFT trained with our Sintel raw data (S\textsuperscript{raw}) outperforms or is on par with prior unsupervised works when compared on the Sintel~\cite{sintel} datasets.
When trained with KITTI raw data (K\textsuperscript{raw}), we are worse than the state of the art and compete with SelFlow~\cite{liu2019selflow} and DDFlow~\cite{liu2019ddflow}.

\paragraph{Comparison with Supervised Methods.}
Since our dataset is synthetic, we can compete and compare with supervised methods trained on Chairs~\cite{dosovitskiy2015flownet} and Things~\cite{flyingthings} and evaluate the capability to generalize on unseen data from Sintel~\cite{sintel} and KITTI~\cite{kitti}.
We show RAFT trained with our data in Table~\ref{tbl:comparison-supervised}.
Following Teed \etal \cite{teed2020raft}, we group methods by the dataset(s) they were trained on. 
We find that RAFT trained on the Chairs and Things (C + T) datasets achieves a better performance than us on Sintel and KITTI except for the outlier ratio F1-all on the KITTI training split where we achieve 16.03\%.
% When comparing the numbers of our ``C + T`` training  with the 
We find that the additional 100k iterations of training from our ``C'' training to our ``C + T`` training does not result in a significant reduction in EPE, hence this shows that RAFT is fitting our data already well in the ``C'' stage and does not benefit as much when training for longer. 
While our dataset combined with the state of the art RAFT yields competitive results, the variability of our data seems to be the major factor of influence. 
% While on the one hand our training labels are generated in an unsupervised way, at the same time we can compete with supervised methods since the only difference is the data. 

We continue the training with the ground truth image- and optical flow data from Sintel~\cite{sintel} and KITTI~\cite{kitti} to compare with other methods in the finetuned regime.
We again follow the training schedule and settings from RAFT~\cite{teed2020raft} with the exception that we do not include the Things dataset and only finetune on either the Sintel or the KITTI dataset.
To prevent overfitting, we only train for 50k iterations in contrast to RAFT-ft which was finetuned for 100k iterations.
Results are shown in Table~\ref{tbl:comparison-supervised}.

\begin{table*}[t]
    % \footnotesize	
    \centering
    \ra{1.3}
    \begin{tabular}{c|lcccccccc}
        \toprule
            \multirow{2}{*}{Data} &
            \multirow{2}{*}{Method} & 
            \multicolumn{2}{c}{SINTEL (train)} & 
            \multicolumn{2}{c}{SINTEL (test)} &
            \multicolumn{2}{c}{KITTI 15 (train)} &
            \multicolumn{1}{c}{KITTI 15 (test)} &
            
        \\
        % \cmidrule{2-3} \cmidrule{5-6} \cmidrule{8-9}
            & & clean & final & clean & final & EPE & F1 [\%] & F1 [\%]
        \\ 
        \midrule
        C & RAFT~\cite{teed2020raft}
        & 2.26 & 4.51 & - & - & 9.85 &  37.56 & -
        \\
        C & RAFT~\cite{teed2020raft} \emph{reproduced}
        & 2.32 & 4.68 & - & - & 10.95 & 40.12 & -
        \\
        \sraw / \kraw & \textbf{RAFT-Ours} ``C''
        & \best{2.00} & \best{3.49} & - & - & \best{6.17} & \best{16.77} & -
        \\
        \kraw / \sraw & \textbf{RAFT-Ours} ``C''
        & 2.60 & 4.15 & - & - & 6.41 & 17.22  & -
        \\
        % \traw & \textbf{RAFT-Ours} ``C''
        % & 2.08 & 3.72 & - & - & - & - & -
        % \\
        
        \midrule
        C + T & LiteFlowNet~\cite{hui18liteflownet}
        & 2.48 & 4.04 & - & - & 10.39 & 28.5 & - 
        \\
        C + T & PWC-Net~\cite{sun2018pwc}
        & 2.55 & 3.93 & - & - & 10.35 & 33.7 & -
        \\
        C + T & LiteFlowNet2~\cite{hui2019lightweight}
        & 2.24 & 3.78 & - & - & 8.97 & 25.9 & -
        \\
        C + T & VCN~\cite{yang2019volumetric}
        & 2.21 & 3.68 & - & - & 8.36 & 25.1 & -
        \\
        C + T & MaskFlowNet~\cite{zhao2020maskflownet}
        & 2.25 & 3.61 & - & - & - & 23.1 & -
        \\
        C + T & FlowNet2~\cite{ilg2017flownet}
        & 2.02 & $3.54^1$ & 3.96 & 6.02 & 10.08 & 30.0 & - 
        \\ 
        C + T & RAFT~\cite{teed2020raft}
        & \best{1.43} & \best{2.71} & - & - & \best{5.04} & 17.4 & -
        \\
        C + T & RAFT~\cite{teed2020raft} \emph{reproduced}
        & 1.52 & 2.68 & - & - & 5.08 & 17.58 & -
        \\
        \sraw / \kraw & \textbf{RAFT-Ours} ``C + T''  
        & \underline{1.93} & \underline{3.47} & 4.06 & 5.97 & \underline{5.99} & \best{16.03} & 15.75
        \\
        \kraw / \sraw & \textbf{RAFT-Ours} ``C + T''  
        & 2.14 & 3.82 & - & - & 6.66 & 17.05 & - 
        \\
        \midrule
        C + T + S/K & FlowNet2-ft~\cite{ilg2017flownet}
        & (1.45) & (2.01) & 4.16 & 5.74 & (2.30) & (6.8) & 11.48
        \\
        C + T + S/K & HD3~\cite{hd3}
        & (1.87) & (1.17) &  4.79 & 4.67 & (1.31) & (4.1) &  6.55
        \\
        C + T + S/K & IRR-PWC~\cite{hur2019iterative}
        & (1.92) &  (2.51) & 3.84 & 4.58 & (1.63) & (5.3) & 7.65
        \\
        C + T + S/K & ScopeFlow~\cite{bar2020scopeflow}
        & - & - & 3.59 & 4.10 & - &  - & 6.82
        \\ 
        C + T + S/K & RAFT-ft~\cite{teed2020raft}
        & (0.77) & (1.20) & \best{2.08} & 3.41 & (0.64) & (1.5) & \best{5.27}
        \\
        \sraw + S / \kraw + K & \textbf{RAFT-Ours-ft} 
        & (0.75) & (1.30) & \underline{2.43} & \best{3.29} & (0.69) & (1.84) & -
        \\
        % \midrule
        % C + \sraw & \textbf{RAFT-Ours} 
        % & 1.94 & 3.40 &  &  &  &  & 
        % \\
        % C + T + \sraw & \textbf{RAFT-Ours} 
        % & 1.96 & 3.42 &  &  &  &  & 
        % \\
        \bottomrule
    \end{tabular}
    \caption{
        \label{tbl:comparison-supervised}
        \textbf{Comparison with supervised methods} on MPI-Sintel~\cite{sintel} and KITTI~\cite{kitti} benchmarks.
        All numbers are AEPE (average end-point-error) except for KITTI where both EPE and outlier ratio F1 is reported.
        Numbers enclosed by parentheses mean that the training is performed on the same dataset.
        The results for entires marked as (-) were not reported by the original authors.
    }
\end{table*}

\paragraph{Cross Dataset Evaluation.}
While the motions in our data are synthetic, the textures come from the dataset we evaluate on, \ie, so far either Sintel or KITTI.
To study the impact of texture in optical flow performance, we take the model trained with our KITTI and evaluate on the Sintel benchmark. 
We do the same also in reverse, \ie, we take the model trained with our Sintel data and evaluate it on the KITTI benchmark dataset. 
The results of these ablations are listed in the rows below our best results in both Table~\ref{tbl:comparison-unsupervised} and~\ref{tbl:comparison-supervised}.
We observe that the EPE in this cross-evaluation is consistently higher. 
Since the distribution of motions is the same between the two datasets and the only real difference is the texture in the data, these results show, to a limited degree, that the texture does in fact have an influence on the ability to generalize on unseen data.
%We argue that
% The relative difference between the errors in this cross-evaluation is within a relatively small 1 pixels. 
% This shows that RAFT is agnostic to textures and that the differences in performance when comparing our method to others are primarily the result of the variability of motions in the training data.

\begin{figure*}
    \centering
    \includegraphics[width=0.33\textwidth]{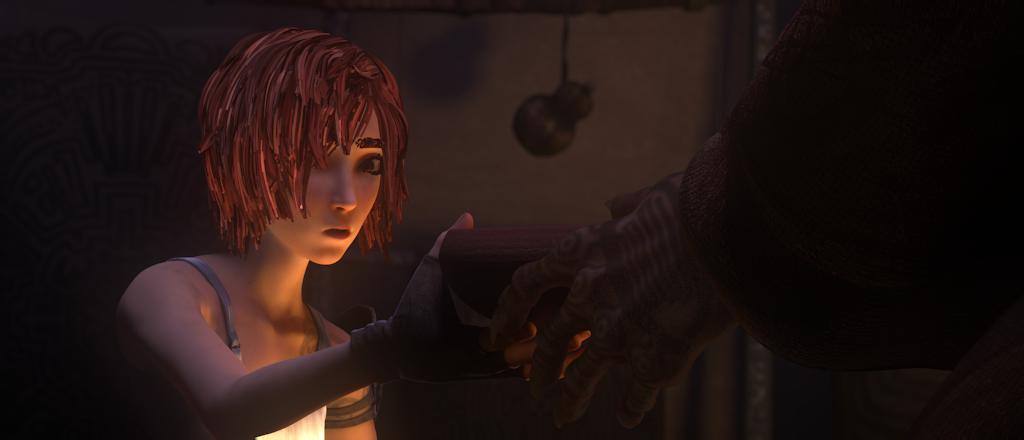}
    \includegraphics[width=0.33\textwidth]{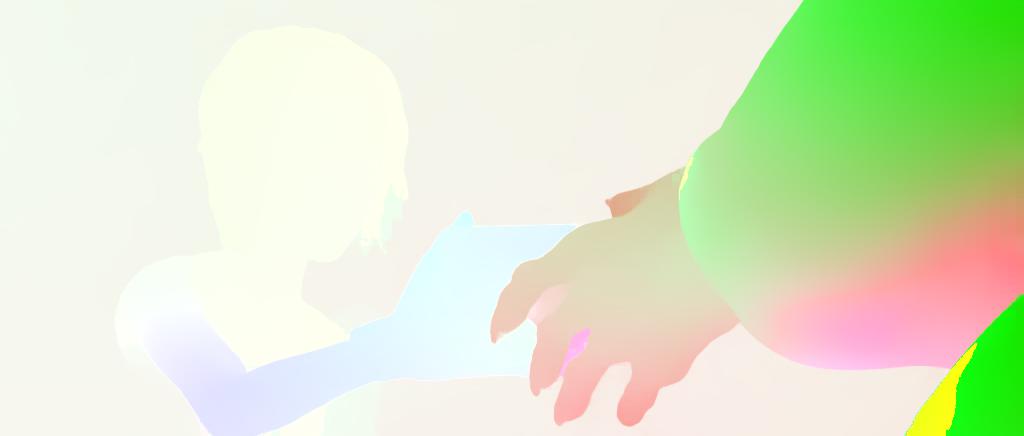}
    \includegraphics[width=0.33\textwidth]{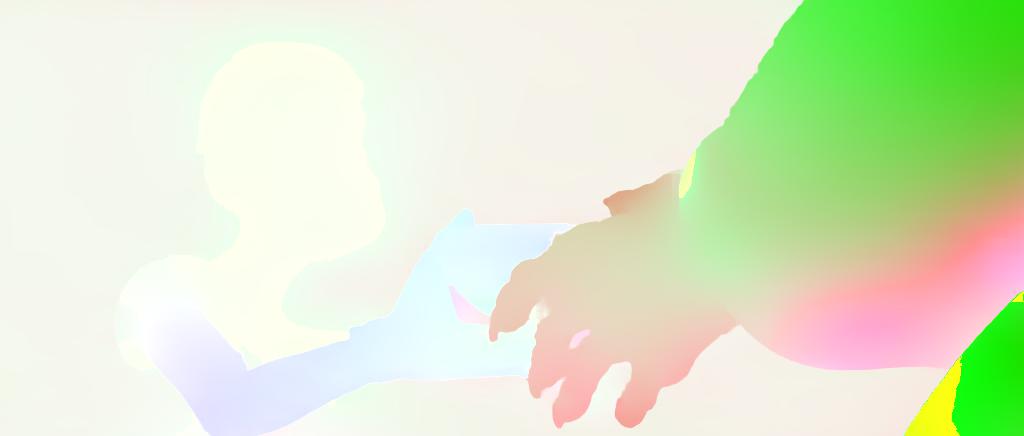}
    \\
    \includegraphics[width=0.33\textwidth]{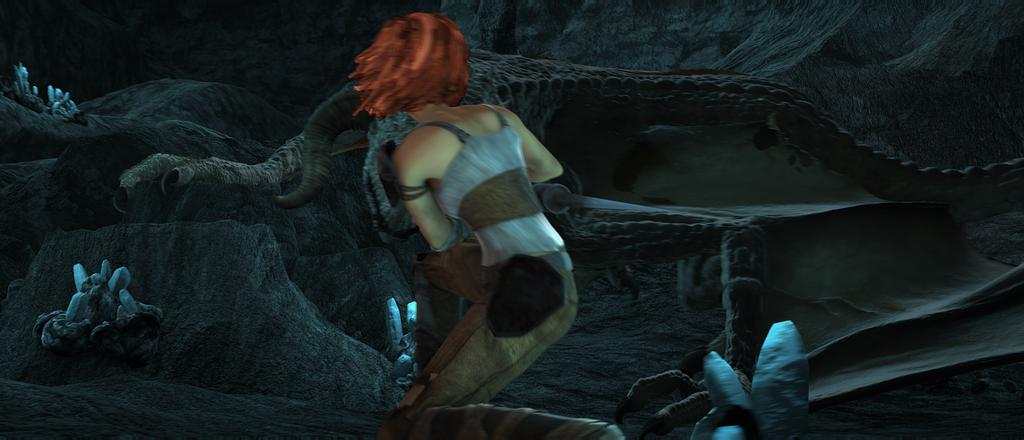}
    \includegraphics[width=0.33\textwidth]{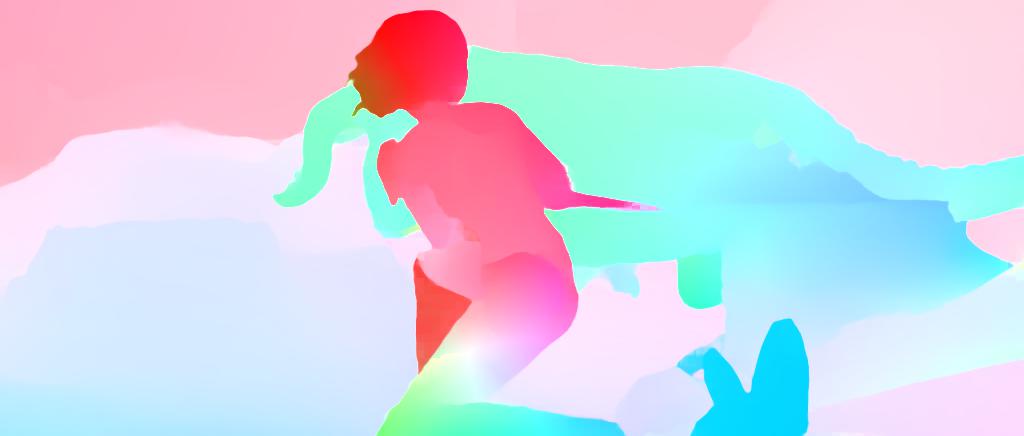}
    \includegraphics[width=0.33\textwidth]{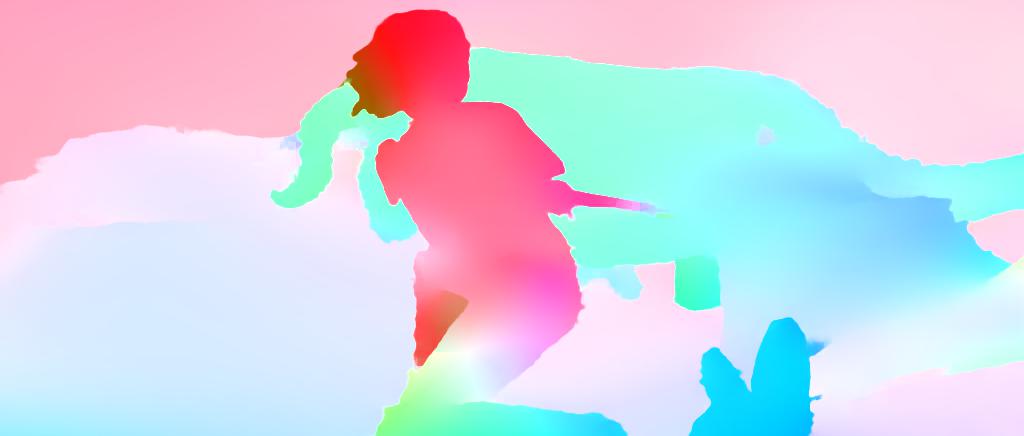}
    \\
    \includegraphics[width=0.33\textwidth]{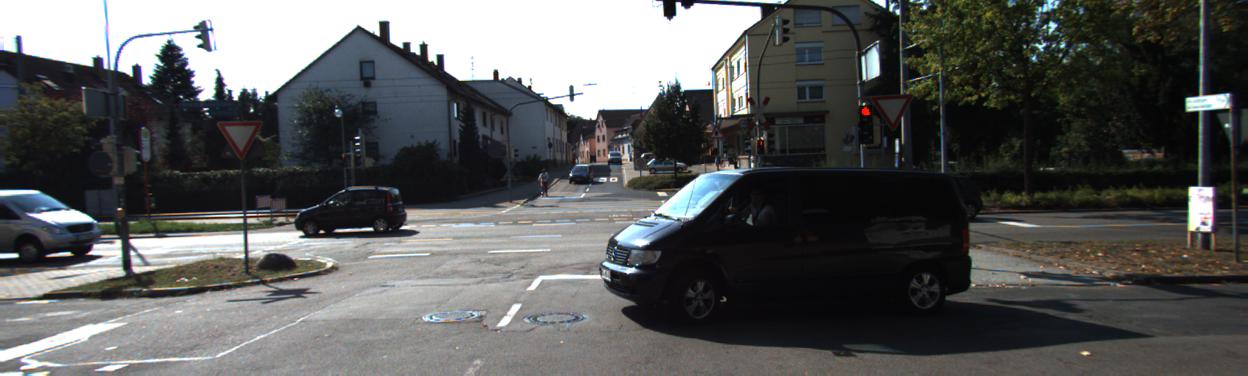}
    \includegraphics[width=0.33\textwidth]{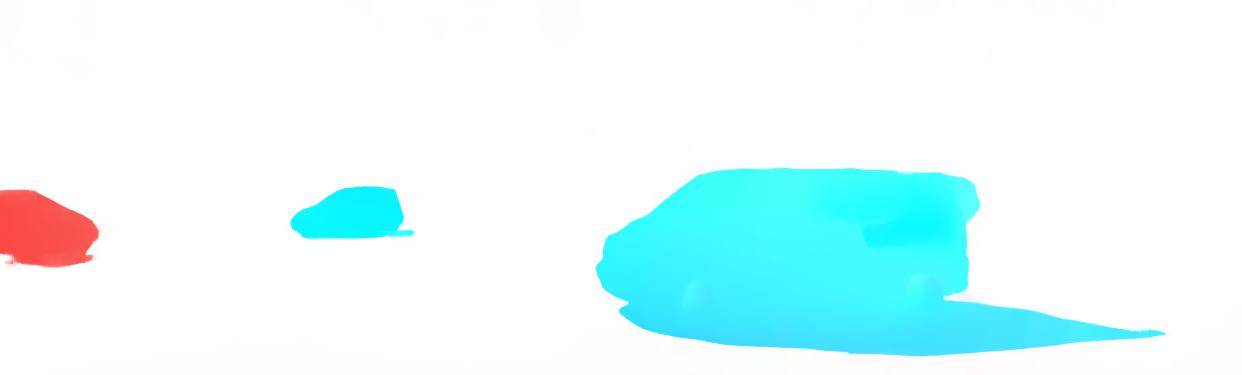}
    \includegraphics[width=0.33\textwidth]{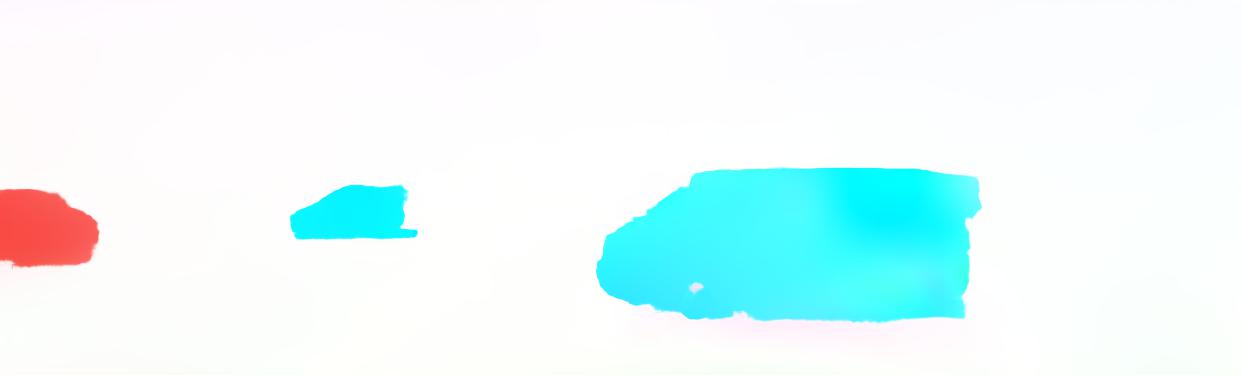}
    % \\
    % \includegraphics[width=0.33\textwidth]{images/outputs/qualitative-comparison/kitti/00172-00-image0.jpg}
    % \includegraphics[width=0.33\textwidth]{images/outputs/qualitative-comparison/kitti/00172-03-their.jpg}
    % \includegraphics[width=0.33\textwidth]{images/outputs/qualitative-comparison/kitti/00172-02-ours.jpg}
    \\
    \includegraphics[width=0.33\textwidth]{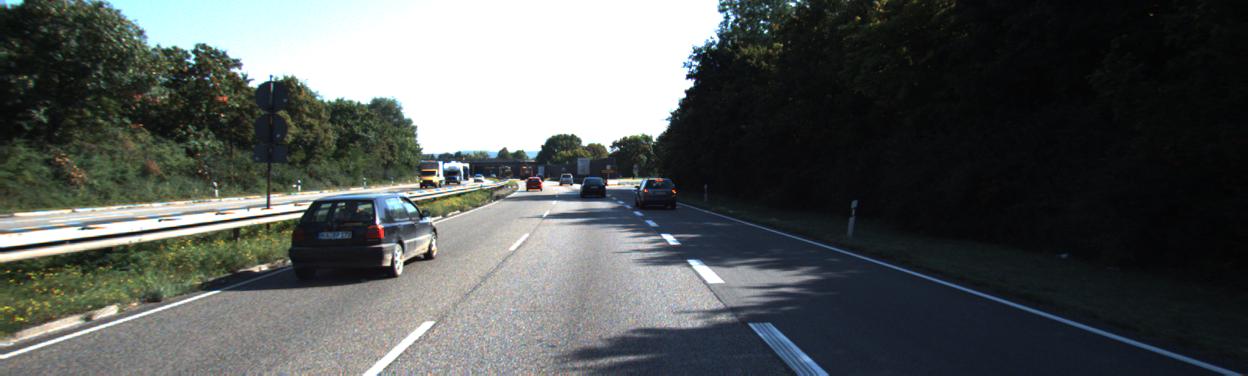}
    \includegraphics[width=0.33\textwidth]{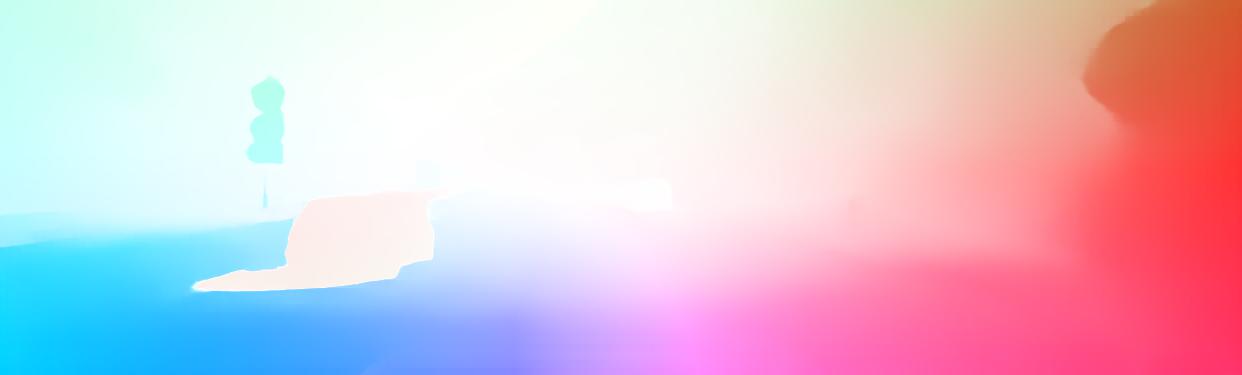}
    \includegraphics[width=0.33\textwidth]{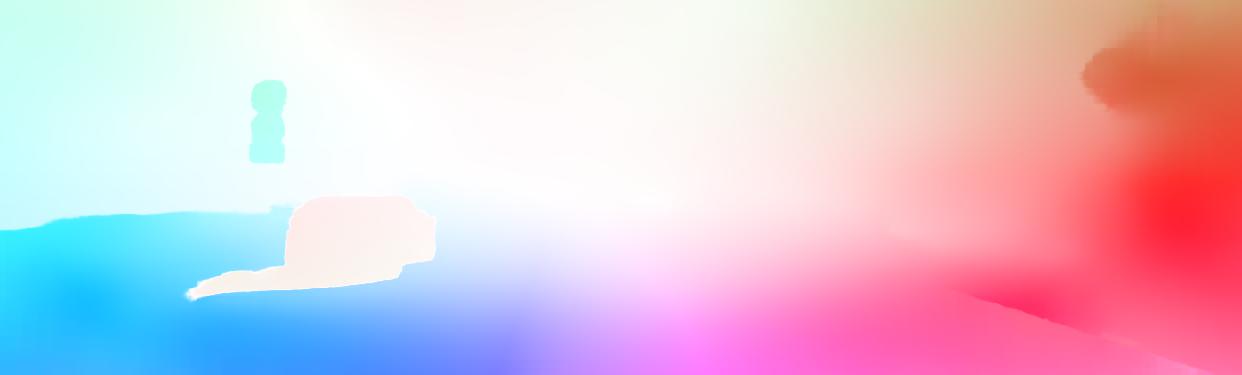}
    \caption{
        \label{fig:qualitative-kitti}
        Qualitative results. We evaluate RAFT~\cite{teed2020raft} trained on C + T (middle) and RAFT trained on our data (right) on the Sintel~\cite{sintel} (top two rows) and KITTI~\cite{kitti} test set (bottom two rows).
    }
\end{figure*}

\paragraph{Qualitative Results.}
In addition to the quantitative evaluations, we also show qualitative flow predictions of RAFT trained with our data vs.\ RAFT trained on Chairs and Things in Figure~\ref{fig:qualitative-kitti}. 
The input images are from the KITTI~\cite{kitti} and Sintel~\cite{sintel} test split and we only show the first image.
Most notably, the flow predictions from our method and also the predictions of RAFT~\cite{teed2020raft} contain the shadow as part of the moving car. 
By the strict definition of optical flow and photometric consistency, this estimate for shadows is correct. 
However, the ground truth data in these regions does not agree and instead assigns the motion of the ground plane to the region where we predict the motion of the shadow.
It is clear that this discrepancy between real optical flow and sensor data from KITTI~\cite{kitti} is a major problem for evaluating the performance of optical flow models as this systematic error cannot be reduced even if a method is qualitatively better in these ambiguous regions.

% KITTI, the missing data, shadows (ambiguous, associated to ground or shadow?) , windows!

\paragraph{Ablation Study.}
To study the effect of the various hyperparameters in the flow synthesis, we generate several smaller datasets of 5k samples each using the MPI-Sintel images.
in Table~\ref{tbl:ablation-main} shows the effect on EPE for different settings of $K$ (number of coarse-to-fine segmentation maps to choose from), $L$ (grid size of TPS warping), $M$ (max.\@ size of a superpixel group in pixels), and $N$ (number of occlusions).
We find that a greater value for $N$ has the highest impact.
Choosing a larger grid size $L$ and using two segmentation maps is beneficial too, however adding a third segmentation map does not improve the EPE further. 
\section{Conclusions and Future Work}

We have presented a novel, unsupervised method to synthesize an unlimited amount of exact ground truth data for optical flow. 
Our approach effectively leverages all available texture in a single image to create realistic occlusions and deformations across several layers of motion.
We compared against supervised as well as unsupervised  methods and showed how our dataset can reduce the reliance on multiple datasets and schedules for pretraining such as FlyingChairs and FlyingThings.
While we made specific choices in regard to the segmentation method or the warping model, our formulation is generic and is compatible with arbitrarily complex segmentation- and motion models.
%Experiments  indicate that more variability is needed to close the domain gap between our synthetic data and the real world. 
We believe that introducing even more variability to the motions in our dataset is a promising direction for future work that could address the generalization gap where we currently fall short or are on par with state of the art methods.

\begin{table}[t]
    % \footnotesize	
    \centering
    \ra{1.3}
    \begin{tabular}{lccc|cc}
        \toprule
        K & L & N & M & clean & final\\
        \midrule
        1 & 4 & 10 & 25k & 2.77 & 4.10\\ 
        \midrule
        1 & 4 & \best{15} & 25k & 2.39 & 3.57\\ 
        1 & 4 & \best{5}  & 25k & 2.61 & 3.85\\ 
        1 & \best{6} & 10 & 25k & 2.58 & 4.00\\ 
        1 & \best{2} & 10 & 25k & 2.65 & 4.31\\ 
        1 & 4 & 10 & \best{50k} & 2.47 & 3.83\\ 
        1 & 4 & 10 & \best{5k}  & 3.18 & 4.35\\ 
        \best{2} & 4 & 10 & 25k & 2.42 & 3.72\\ 
        \best{3} & 4 & 10 & 25k & 2.44 & 3.72\\ 
        \bottomrule 
    \end{tabular}
    \caption{
        \label{tbl:ablation-main}
        \textbf{Impact of dataset hyperparameters.} 
        All experiments use textures from MPI-Sintel ~\cite{sintel}.
        All numbers are AEPE (average end-point-error).
        Parameters in bold mark the difference to the baseline (first row).
    }
\end{table}

%\nocite{*}

{\small
\bibliographystyle{ieee_fullname}
\bibliography{main}
}

\end{document}